\documentclass[conference,letterpaper]{IEEEtran}
\IEEEoverridecommandlockouts
\usepackage[dvips]{graphicx}
\usepackage{multirow}
\usepackage[left=0.71in,top=0.94in,right=0.71in,bottom=1.18in]{geometry}
\usepackage{amssymb}  \usepackage{amsmath}
\DeclareMathOperator*{\argmax}{argmax}

\DeclareMathOperator*{\st}{s.t.}

\DeclareMathOperator*{\avg}{avg}
\usepackage[ruled,vlined,linesnumbered]{algorithm2e}
\SetAlFnt{\scriptsize}
\SetAlCapFnt{\footnotesize}
\SetAlCapNameFnt{\footnotesize}
\usepackage{stfloats}
\usepackage{graphicx}
\usepackage[caption=false, font=footnotesize]{subfig}
\usepackage{bm}

\makeatletter
\let\NAT@parse\undefined
\makeatother

\makeatletter
\newcommand{\removelatexerror}{\let\@latex@error\@gobble}
\makeatother

\usepackage{xcolor}

\begin{document}
\title{\huge End-to-end Decentralized Multi-robot Navigation in Unknown Complex Environments via Deep Reinforcement Learning}

\author{\authorblockN{Juntong Lin, Xuyun Yang, Peiwei Zheng and Hui Cheng$^{*}$}
\authorblockA{\textit{School of Data and Computer Science}\\
\textit{Sun Yat-sen University}\\
\textit{Guangzhou, Guangdong Province, China}\\
$^{*}$Corresponding author: chengh9@mail.sysu.edu.cn
}}

\maketitle

\begin{abstract}
In this paper, a novel deep reinforcement learning (DRL)-based method is proposed to navigate the robot team through unknown complex environments, where the geometric centroid of the robot team aims to reach the goal position while avoiding collisions and maintaining connectivity. Decentralized robot-level policies are derived using a mechanism of centralized learning and decentralized executing. The proposed method can derive end-to-end policies, which map raw lidar measurements into velocity control commands of robots without the necessity of constructing obstacle maps. Simulation and indoor real-world unmanned ground vehicles (UGVs) experimental results verify the effectiveness of the proposed method.
\end{abstract}


 \section{INTRODUCTION}

Multi-robot system has broad applications such as search, rescue, forest inspection, agricultural spraying, collaborative transportation, etc. When a team of robots collaboratively accomplishes tasks in complex environments, multi-robot navigation techniques are essential to guarantee the safety of the robot team \cite{1570754}. When the workspaces of the robot team are unknown and complex, the navigation policies are required to be both on-line and able to handle obstacles of different shapes. In addition, the communication range of each robot is usually constrained, thus the navigation policy should take connectivity into consideration to ensure communication among the robots. Furthermore, decentralized navigation strategies are preferable to centralized ones on enabling autonomous decision making for each robot.

\begin{figure}[]
  \centering
  \includegraphics[width=0.99\linewidth]{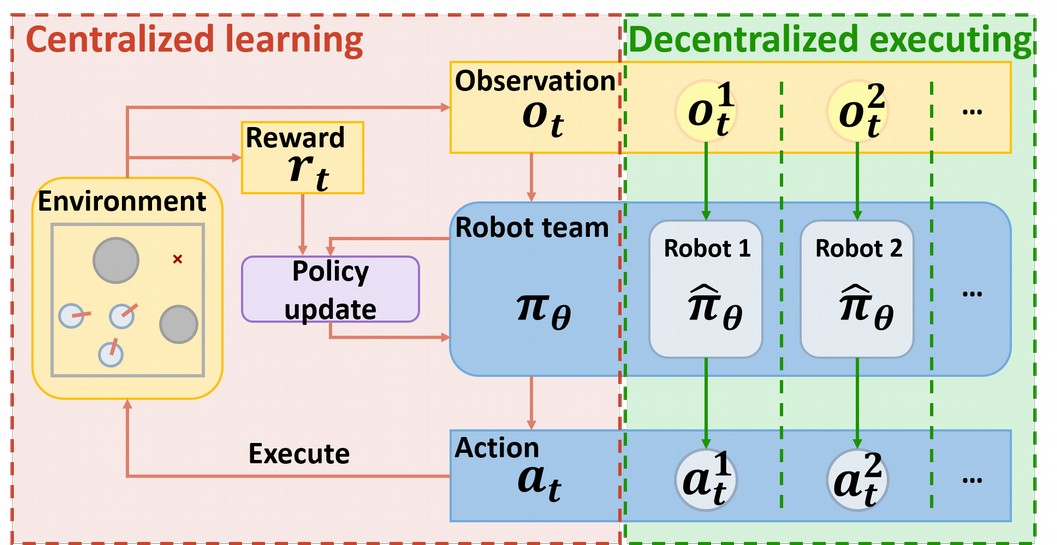}
  \vspace{-15pt}
  \caption{An overview of our method. During centralized learning period, the robot team interacts with the environment and evolves a team-level policy $\pi_{\bm{\theta}}$ in a trial-and-error manner. In our method, a robot-level policy $\hat{\pi}_{\bm{\theta}}$ can be determined once the team-level policy $\pi_{\bm{\theta}}$ is obtained. With the robot-level policy $\hat{\pi}_{\bm{\theta}}$, each robot can make decision merely depending on local observations in decentralized executing stage. }
  \label{framework}
  \vspace{-20pt}
\end{figure}

This paper studies the decentralized multi-robot navigation in unknown complex environments as shown in Fig. \ref{adaptive_basic} and Fig. \ref{adaptive_valid}, where the geometric centroid of the formation is required to reach the goal position efficiently while avoiding collision and maintaining connectivity of the robot team. A deep reinforcement learning (DRL)-based approach is proposed to accomplish the multi-robot navigation task. By means of a centralized learning and decentralized executing mechanism as shown in Fig. \ref{framework}, a decentralized end-to-end robot-level policy can be obtained, which directly maps raw lidar range data to continuous velocity control commands without the necessity of constructing obstacle maps. 

\subsection{Related work}

\begin{figure} 
  \centering
%  \hfill
  \subfloat[Scene description.]
  {  \includegraphics[width=0.31\linewidth]{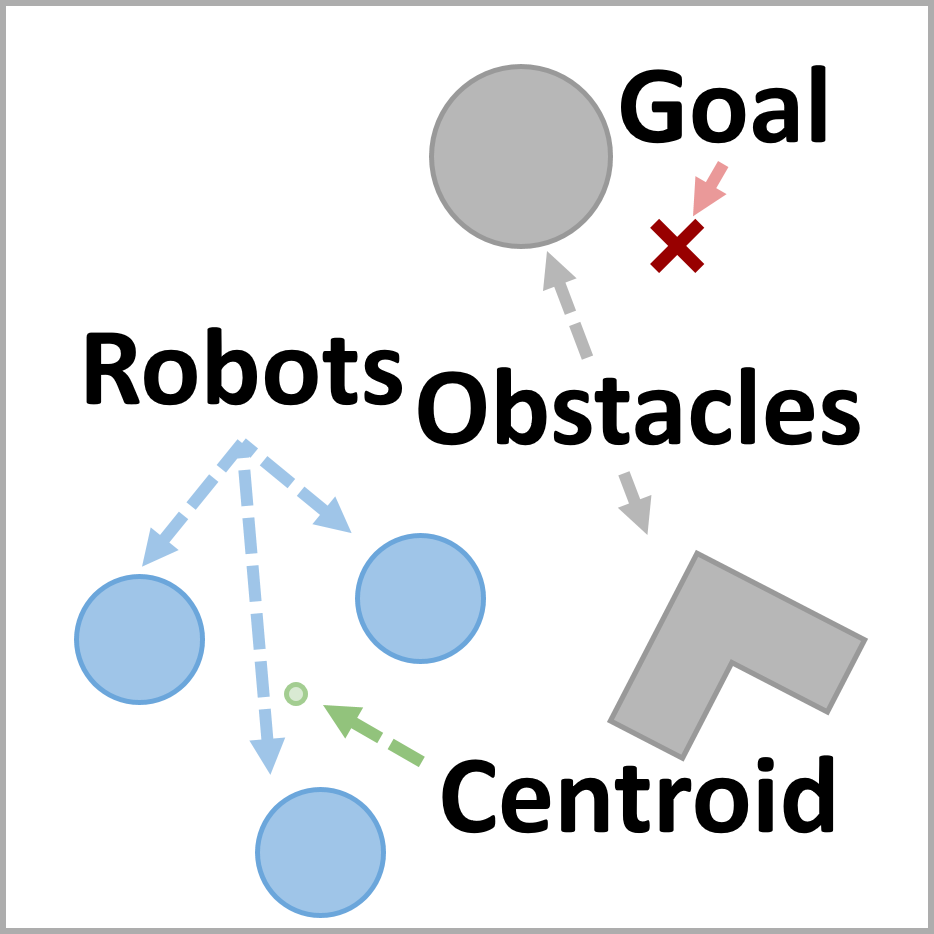}
  \label{adaptive_basic}
  }
%  \hfill
  \subfloat[Adaptive control.]
  {  \includegraphics[width=0.31\linewidth]{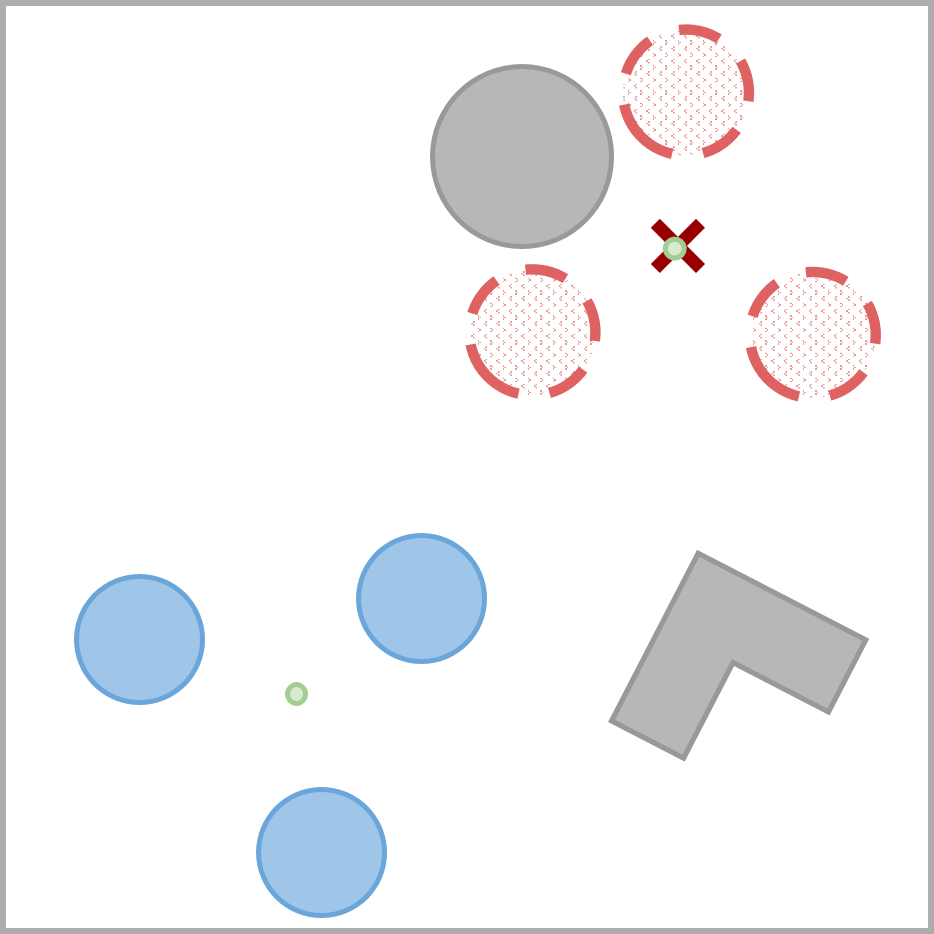}
  \label{adaptive_valid}
  }
%  \hfill
  \subfloat[Invlid assignment.]
  {  \includegraphics[width=0.31\linewidth]{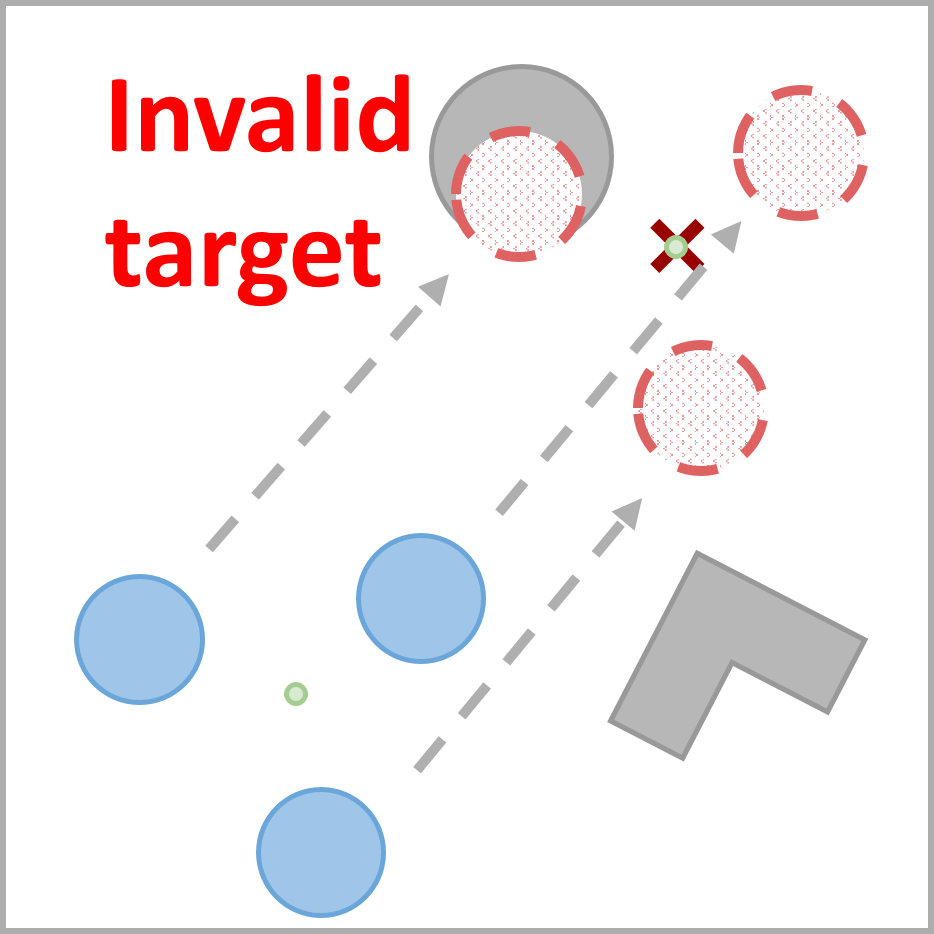}
  \label{adaptive invalid}
  }
%  \hfill
  \caption{Multi-robot navigation task.}
  \label{adaptive} 
  \vspace{-12pt}
\end{figure}
 
There exists extensive research work for multi-robot navigation, which can be further categorized into  rule-based and learning-based strategies. Rule-based approaches include using leader-follower scheme \cite{koo2001formation}, artificial potential field (APF) \cite{balch2000social}, graph theory \cite{wasik2016graph}, consensus theory \cite{1605401}, model predictive control \cite{doi:10.1177/0278364914530482}, virtual structure \cite{alonso2015multi, alonso2016distributed}, etc. 

In the rule-based navigation methods, the obstacle map should be constructed using the sensor data. The performance of the rule-based methods highly relies on the constructed obstacle map. In addition, the real-time mapping of the navigation environment with the robots' onboard sensors is challenging and computationally prohibitive sometimes.

Learning-based methods are alternatives for rule-based methods. The learning-based approaches can derive end-to-end policies which map raw sensor data to control commands without the necessity of constructing obstacle maps. 

Most of the existing learning-based navigation methods focus on single-robot settings \cite{duguleana2016neural, xie2017towards, kahn2017self}. In the case of multi-robot systems, the research work mainly focuses on local collision avoidance \cite{long2017deep, long2017towards, chen2017socially}, where multiple robots move to their designated goal positions correspondingly without colliding with other robots and the obstacles. As shown in Fig. \ref{adaptive invalid}, it is difficult to assign the goal position to each robot beforehand when the robot team navigates through unknown cluttered environments. Moreover, as the communication range of the robot is usually constrained, the information exchange among the robots cannot be ensured if the connectivity of the robot team is not considered. 

In this paper, a DRL-based method is proposed to accomplish the multi-robot navigation task, where the geometric centroid of the robot team aims to reach the goal position while avoiding collisions and maintaining connectivity.

\subsection{Contribution and paper organization}

In this paper, the multi-robot navigation problem is studied. The main contributions of the paper are presented as follows:

\begin{itemize}
\item A novel DRL-based method is proposed to navigate the multi-robot team through unknown cluttered environments in a decentralized manner. The derived policy directly maps the raw lidar measurements into the continuous velocity control commands of the robots without the necessity of constructing the obstacle maps.
\item Simulation results illustrate that the robot team can navigate through unknown complex environments based on the proposed method. Comparison results show that our proposed method (which makes decisions based on local observations) outperforms an APF-based method (which can obtain global environment information) in narrow scenarios.
\item A team of 3 holonomic unmanned ground vehicles (UGVs) is built to serve as a real-world policy verification platform. Indoor real-world experimental results show that the learned policy can successfully navigate a team of 3 holonomic UGVs through the dense obstacles.
\end{itemize}

The remainder of this paper is organized as follows: In Section \ref{sec:formulation}, the multi-robot navigation problem is described, and the problem formulation is addressed. The proposed DRL-based navigation method is presented in Section \ref{sec:method}. In Section \ref{sec:experiment}, numerical and experimental results are presented to illustrate its effectiveness. Section \ref{sec:conclusion} concludes the paper.
 \section{PROBLEM FORMULATION}
\label{sec:formulation}

In this section, the multi-robot navigation problem is formulated as a partially observable Markov decision process (POMDP).

\begin{figure} 
  \centering
  \subfloat[Local observation.]
  {  \includegraphics[width=0.4\linewidth]{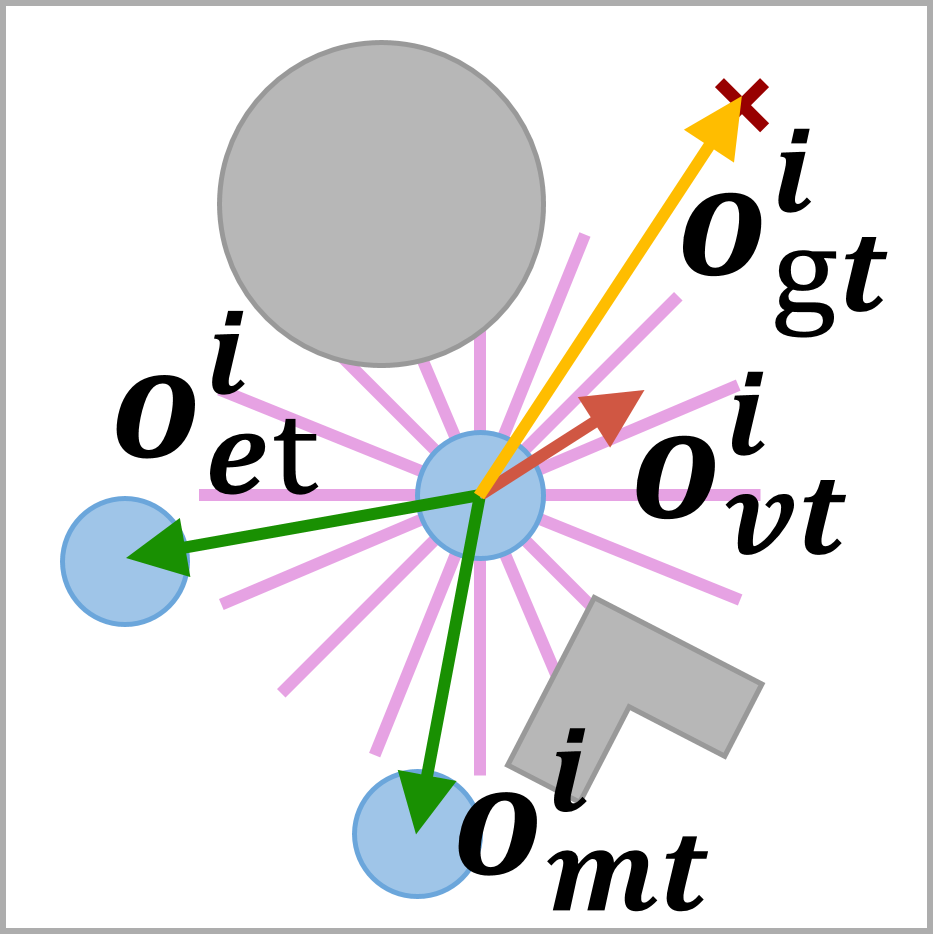}
  \label{lidar}
  }
  \subfloat[Local frame.]
  {  \includegraphics[width=0.4\linewidth]{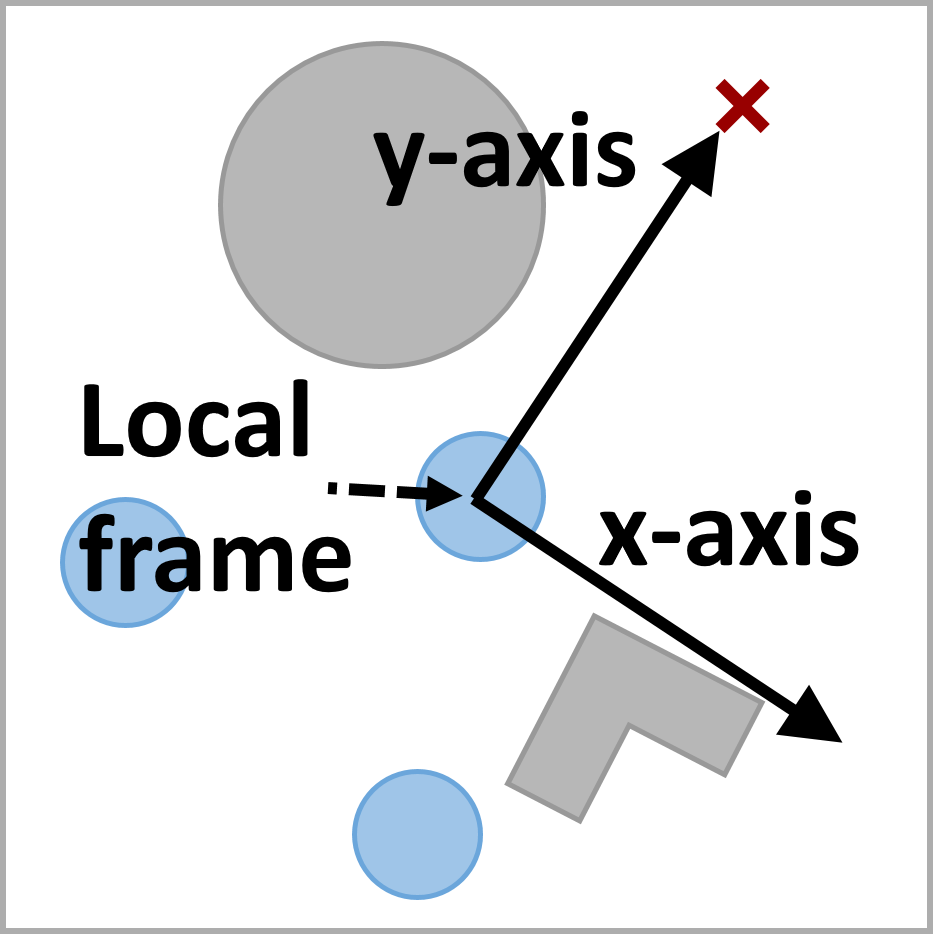}
  \label{local frame} 
  }
  \hfill
  \caption{Setting description. Information about teammates $\bm{o}^i_{mt}$, goal $\bm{o}^i_{gt}$ and current velocity $\bm{o}^i_{vt}$ are green, yellow and red arrows, respectively. Information about environment $\bm{o}^i_{et}$ are purple lines. Vectors are represented under robots' local frames.}
  \label{Problem description} 
  \vspace{-20pt}
\end{figure}
 
In this paper, a team of homogeneous robots is considered, where the robots are holonomic with equal radius $R$. Assume there are $N$ robots and $M$ static obstacles in a navigation environment. The obstacles are modeled as polygons, and the $j$-th obstacles ($1 \leq j \leq M$) is denoted as $\tilde{\bm{p}}^j = [\tilde{\bm{p}}^j_1, \tilde{\bm{p}}^j_2, \dots, \tilde{\bm{p}}^j_{z_j}]$, where $\tilde{\bm{p}}^j_{z_j}$ is the position of the $z_j$-th vertex of the $j$-th obstacles and $z_j$ is the number of vertices of the $j$-th obstacle. At each time step $t$, The $i$-th robot ($1 \leq i \leq N$) with velocity $\bm{v}^i_t$ locates at $\bm{p}_t^i$. The robot team computes the control command $\bm{a}_t = [\bm{a}_t^1, \bm{a}_t^2, \dots , \bm{a}_t^N]$ based on the local observations of all robots $\bm{o}_t = [\bm{o}_t^1, \bm{o}_t^2, \dots , \bm{o}_t^N]$ by sampling from the distribution $\pi_{\bm{\theta}}(\bm{o}_t)$, where the distribution $\pi_{\bm{\theta}}(\bm{o}_t)$ is generated from the policy $\pi_{\bm{\theta}}$ and $\bm{\theta}$ denotes the policy parameters. Within the time horizon $\Delta t$, the positions of robots are updated according to the control value $\bm{a}_t$.

The local observation $\bm{o}^i_t$ of the $i$-th robot consists of current velocity $\bm{o}^i_{vt}$, relative positions of robot team's goal $\bm{o}^i_{gt}$ as well as other $N-1$ robots $\bm{o}^i_{mt}$, and the partial environment information $\bm{o}^i_{et}$. We assume the local observation is always obtainable for the policy during the navigation. The form of the partial environment information $\bm{o}^i_{et}$ and the control value $\bm{a}^i_t$ can be various depending on the configuration of the multi-robot system. In this paper, we study the case where robots perceive the environment with 2D laser scanners and take velocity as control commands. Therefore, the partial environment information $\bm{o}^i_{et}$ represents the raw 2D laser measurements of the $i$-th robot as shown in Fig. \ref{lidar} and $\bm{a}^i_t$ denotes the velocity control command. Vectors (e.g., velocity control command $\bm{a}^i_t$, current velocity $\bm{o}^i_{vt}$, relative position of the goal $\bm{o}^i_{gt}$, and relative positions of other robots $\bm{o}^i_{mt}$) are represented under the robots' local frame as shown in Fig. \ref{local frame}.

Accordingly, the multi-robot navigation problem can be formulated as a POMDP. A POMDP can be denoted as a 6-tuple $(\mathcal{S}, \mathcal{A}, \mathcal{O}, \mathcal{T}, \mathcal{\varepsilon}, \mathcal{R})$, where $\mathcal{S}$ is the state space, $\mathcal{A}$ is the action space, $\mathcal{O}$ is the observation space, $\mathcal{T}$ is the state transition distribution, $\mathcal{\varepsilon}$ is the emission distribution, and $\mathcal{R}$ is the reward function.

Specifically, the vector $\bm{s}_t = [\bm{p}_t, \bm{v}_t, R, \tilde{\bm{p}}, \bm{g}] \in \mathcal{S}$ represents the states of all the robots at the time step $t$, where $\bm{p}_t = [\bm{p}_t^1, \bm{p}_t^2, \dots , \bm{p}_t^N]/\bm{v}_t = [\bm{v}_t^1, \bm{v}_t^2, \dots , \bm{v}_t^N]$ denote the positions/velocities of all the $N$ robots, and $\tilde{\bm{p}} = [\tilde{\bm{p}}^1, \tilde{\bm{p}}^2, \dots , \tilde{\bm{p}}^M]$ denotes all the $M$ obstacles and $\bm{g}$ denotes the goal position. The vector $\bm{a}_t \in \mathcal{A}$ and $\bm{o}_t \in \mathcal{O}$ denote the actions and observations of all the robots. State transition distribution $\mathcal{T}$ determines $p(\bm{s}_{t+1} \vert \bm{s}_t , \bm{a}_t)$, which is the distribution of the next states $\bm{s}_{t+1}$ on the condition of taking the action $\bm{a}_t$ under the state $\bm{s}_t$. Emission distribution $\mathcal{\varepsilon}$ determines $p(\bm{o}_t \vert \bm{s}_t)$, which is the distribution of the accessible observations of the robots under the state $\bm{s}_t$. Note that both state transition distribution $\mathcal{T}$ and emission distribution $\mathcal{\varepsilon}$ are defined implicitly by the physical meaning of $\bm{s}_t, \bm{a}_t$ and $\bm{o}_t$. The reward function $\mathcal{R}$ is a map $\mathcal{S} \times \mathcal{A} \rightarrow \mathbb{R}$, which returns the scalar immediate reward $r_t$ of taking the action $\bm{a}_t$ under the state $\bm{s}_t$. The design of the reward function $\mathcal{R}(\bm{s}_t, \bm{a}_t)$ is elaborated in Section \ref{sec:method}.

Once formulating the multi-robot navigation problem as a POMDP, we aim to find the optimal parameter $\bm{\theta}^*$ for the policy $\pi_{\bm{\theta}}$ to maximize the expected accumulated reward $\mathbb{E}_{\bm{\tau} \sim p_{\pi_{\bm{\theta}}}(\bm{\tau})}[\sum_{t=1}^T \mathcal{R}(\bm{s}_t, \bm{a}_t)]$, where $T$ refers to the maximum time step, $\tau = [\bm{s}_1, \bm{o}_1, \bm{a}_1, \dots , \bm{s}_T, \bm{o}_T, \bm{a}_T]$ is the trajectory of the robot team, $p_{\pi_{\bm{\theta}}}(\bm{\tau})$ denotes the distribution of a trajectory $\bm{\tau}$ under policy $\pi_{\bm{\theta}}$, state transition distribution $\mathcal{T}$ and emission distribution $\mathcal{\varepsilon}$. The finding of the optimal parameter $\bm{\theta}^*$ can be formulated as:
\begin{align}
\begin{split}
\argmax_{\bm{\theta}} & \quad \mathbb{E}_{\bm{\tau} \sim p_{\pi_{\bm{\theta}}}(\bm{\tau})}[\sum_{t=1}^T \mathcal{R}(\bm{s}_t, \bm{a}_t)] \\
\st & \quad p_{\pi_{\bm{\theta}}}(\bm{\tau}) = p(\bm{s}_1) p(\bm{o}_1 \vert \bm{s}_1) \pi_{\bm{\theta}}(\bm{a}_1 \vert \bm{o}_1) \\ 
& \hspace{-10pt} \prod_{t=1}^{T-1} [p(\bm{s}_{t+1} \vert \bm{s}_t, \bm{a}_t) p(\bm{o}_{t+1} \vert \bm{s}_{t+1}) \pi_{\bm{\theta}}(\bm{a}_{t+1} \vert \bm{o}_{t+1}) ] \\
\end{split}
\label{opt}
\end{align} \section{METHODOLOGY}
\label{sec:method}

In this section, we first present a DRL-based method to derive a centralized policy, where the whole robot team is regarded as an entity. Then the method is adapted to derive a decentralized policy to enhance the autonomy of each robot.

\subsection{Centralized policy learning}
\label{subsec:centralized}

In this part, the robot team is regarded as an entity. To solve the POMDP defined in \ref{sec:formulation}, the reward function $\mathcal{R}(\bm{s}_t,\bm{a}_t)$ is presented, which primarily rewards desirable robot team behaviors and penalizes unwanted manners. It can be formulated as:
\begin{align}
\begin{split}
\mathcal{R}(\bm{s}_t,\bm{a}_t) = & r_e + w_g\mathcal{R}_{g}(\bm{s}_t, \bm{a}_t) + w_c\mathcal{R}_{c}(\bm{s}_t, \bm{a}_t) \\
&+ w_f\mathcal{R}_{f}(\bm{s}_t, \bm{a}_t) + w_p\mathcal{R}_{p}(\bm{s}_t, \bm{a}_t) ,
\end{split}
\label{reward_function}
\end{align}
where $r_{e}$ is a negative constant serving as a penalty to motivate the robot team to reach the goal position as fast as possible; the reward functions $\mathcal{R}_{g}(\bm{s}_t, \bm{a}_t)$, $\mathcal{R}_{c}(\bm{s}_t, \bm{a}_t)$, $\mathcal{R}_{f}(\bm{s}_t, \bm{a}_t)$ and $\mathcal{R}_{p}(\bm{s}_t, \bm{a}_t)$ are related to the objectives of reaching the goal position, avoiding collisions, maintaining connectivity, and achieving smooth motion of the robots, respectively; $w_g, w_c, w_f$ and $w_p$ are the corresponding weights. The reward functions are addressed in detail as follows.

The robot team will be rewarded by $r_{goal}$ when the geometric centroid $\bar{\bm{p}}$ of the robot team reaches the goal position $\bm{g}$, and get a reward or penalty depending on the centroid $\bar{\bm{p}}$ moving to or away from the goal. The reward function $\mathcal{R}_{g}(\bm{s}_t, \bm{a}_t)$ can be formulated as:
\begin{align}
\begin{split}
	&\mathcal{R}_{g}(\bm{s}_t, \bm{a}_t)\\
	&=\begin{cases}
	r_{goal}				& \hspace{-12pt} \text{if $ \| \bar{\bm{p}}_{t+1} - \bm{g} \|_2 \leq \epsilon_g$},\\
	\begin{aligned}
	&\| \bar{\bm{p}}_t - \bm{g} \|_2 -
	\| \bar{\bm{p}}_{t+1} - \bm{g} \|_2
	\end{aligned}              & \hspace{-12pt} \text{otherwise},
	\end{cases}
\end{split}
\end{align}
where $\epsilon_g$ refers to the radius of goal zone.

When any robot collides with other robots or obstacles in the environment, the robot team will be penalized: 
\begin{equation}
\mathcal{R}_{c}(\bm{s}_t, \bm{a}_t)=
\begin{cases}
r_{collision} 	
& 
\begin{aligned} 
&\text{if } \bm{p}_{t+1}^i \text{intersects with } \tilde{\bm{p}}^j  \\
&\text{or } \| \bm{p}_{t+1}^i - \bm{p}_{t+1}^k \|_2 < 2R \\
&\text{for any } i, j, k (i \neq k)
\end{aligned} \\
0 & \text{otherwise}.
\end{cases}
\end{equation}
where $r_{collision}$ is a negative value.

The robot team will get a penalty if the connectivity is not guaranteed (distances among the robots exceeding the communication range threshold $d$):
\begin{equation}
\mathcal{R}_{f}(\bm{s}_t, \bm{a}_t) = \avg(\{\min(0, d - \| \bm{p}_{t+1}^i - \bm{p}_{t+1}^j  \|_2) \ \vert \ \forall i \neq j\}),
\end{equation}
where $\avg(\cdot)$ computes the average of a set of values.

Smooth motion of the robot team is desirable, and a penalty is applied with respect to the velocity variation:
\begin{equation}
\mathcal{R}_{p}(\bm{s}_t, \bm{a}_t) = \avg(\{- \arccos \frac{< \bm{v}_t^i, \bm{v}_{t+1}^i>}{\| \bm{v}_t^i \|_2 \| \bm{v}_{t+1}^i \|_2} \ \vert \ \forall i \}),
\end{equation}
where $< \bm{a}, \bm{b} >$ denotes inner product of two vectors.

With the reward functions defined above, actor-critic-based DRL algorithms \cite{sutton1998reinforcement} can be applied to solve the optimization problem in (\ref{opt}). In actor critic-based DRL algorithms, a critic is used to guide the update of actor (i.e., policy $\pi_{\bm{\theta}}$). The critic is based on the value function $V_{\bm{\phi}}$ with parameter $\bm{\phi}$. The value function $V_{\bm{\phi}}$ is a map $\mathcal{O} \rightarrow \mathbb{R}$, which returns the expected accumulated reward under an observation $\bm{o}_t$. Once the form of parametric function aproximators for the policy $\pi_{\bm{\theta}}$ and the value function $V_{\bm{\phi}}$ are determined, actor critic-based DRL algorithms can then be applied to derive a centralized policy $\pi_{\bm{\theta}}$.

\subsection{Decentralized policy learning}

Though a policy $\pi_{\bm{\theta}}$ can be derived with the method mentioned above, the policy $\pi_{\bm{\theta}}$ cannot be applied in a decentralized manner since it takes the global observation of the robot team $\bm{o}_t$ as input, which prohibits the decentralized execution. There exists literatures \cite{tan1993multi, NIPS2017_7217} about deriving decentralized policies for multi-robot tasks, which update the policy for each robot independently. Considering the demand of group behaviors in the navigation task, we adopt a mechanism to obtain the decentralized robot-level policy while updating the joint policy of the robot team during training. Similar mechanism has also succeeded in solving multi-agent graphical games \cite{Zhang2018ModelFreeRL}.

Specifically, decentralized policies are derived by adding a constraint to the policy $\pi_{\bm{\theta}}$ and carrying out the centralized policy learning in Section \ref{subsec:centralized}. Note that the local observation $\bm{o}^i_t$ is sufficient for the $i$-th robot to identify itself, since the robot can infer its position in the team from the relative positions of other $N-1$ robots $\bm{o}^i_{mt}$. It implies that all robots can share a robot-level policy $\hat{\pi}_{\bm{\theta}}$ to make decision as long as the capacity of the function approximator for $\hat{\pi}_{\bm{\theta}}$ is high enough. In other words, the team-level policy $\pi_{\bm{\theta}}$ can be constructed with the robot-level policy $\hat{\pi}_{\bm{\theta}}$. In particular, each robot computes its' action $\bm{a}^i_t$ with the robot-level policy $\hat{\pi}_{\bm{\theta}}$ based on its' local observation $\bm{o}^i_t$. Then the team-level action is obtained by concatenating actions of all robots. The process above can be formulated as a constraint to the team-level policy $\pi_{\bm{\theta}}(\bm{o}_t) = [\hat{\pi}_{\bm{\theta}}(\bm{o}_t^1), \hat{\pi}_{\bm{\theta}}(\bm{o}_t^2), \dots, \hat{\pi}_{\bm{\theta}}(\bm{o}_t^N)]$, where the parameter $\bm{\theta}$ is shared by all robot-level policies. By adding this constraint, a robot-level policy $\hat{\pi}_{\bm{\theta}}$ can be obtained once we find the team-level policy $\pi_{\bm{\theta}}$ by the centralized learning in Section \ref{subsec:centralized}. Hence, the robot team can navigate in a decentralized manner with the robot-level policy $\hat{\pi}_{\bm{\theta}}$ during the execution.

\begin{figure} 
  \centering
  \includegraphics[width=0.99\linewidth]{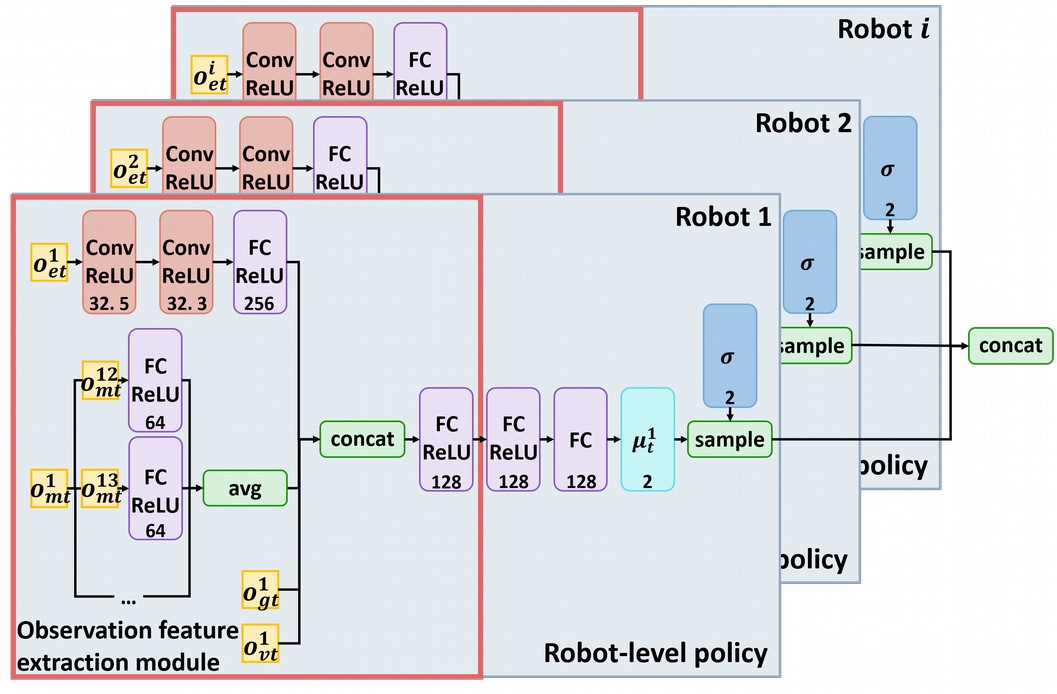}
  \vspace{-20pt}
  \caption{Policy network architecture.}
  \label{policy}
  \vspace{-10pt}
\end{figure}

Considering the high capacity requirements, a neural network is used as the function approximator for the robot-level policy $\hat{\pi}_{\bm{\theta}}$. As shown in Fig. \ref{policy}, the team-level policy $\pi_{\bm{\theta}}$ is constructed with a robot-level policy $\hat{\pi}_{\bm{\theta}}$ shared by all robots. In the robot-level policy $\hat{\pi}_{\bm{\theta}}$, the local observations of each robot sequentially go through an observation feature extraction module (wrapped by red rectangles in Fig. \ref{policy}) and subsequent modules. In the observation feature extraction module, two one-dimensional convolutional layers \cite{fukushima1982neocognitron, lecun1989backpropagation} with ReLU (denoted by Conv-ReLU) are adopted so as to process raw sensor data $\bm{o}^i_{et}$ effectively. Since we consider homogeneous robots, the policy function should be insensitive to the order of relative positions in the teammate observation $\bm{o}^i_{mt}$. Therefore, we process the teammate observation $\bm{o}^i_{mt}$ with an order insensitive submodule. In particular, all relative positions $\bm{o}^{ik}_{mt}$ ($1 \leq k \leq N, k \neq i$) in $\bm{o}^i_{mt}$ are separately processed by a shared fully connected layer with ReLU (denoted by FC-ReLU) and the outputs are averaged subsequently. During execution, each robot samples its action from a Gaussian distribution $\mathcal{N}({\bm{\mu}}^i_t, \bm{\sigma})$, where the mean ${\bm{\mu}}^i_t$ is computed by robot-level policy $\hat{\pi}_{\bm{\theta}}$ based on robot's local observation $\bm{o}^i_t$ and the variance $\bm{\sigma}$ is obtained during training.

\begin{figure} 
  \centering
  \includegraphics[width=0.99\linewidth]{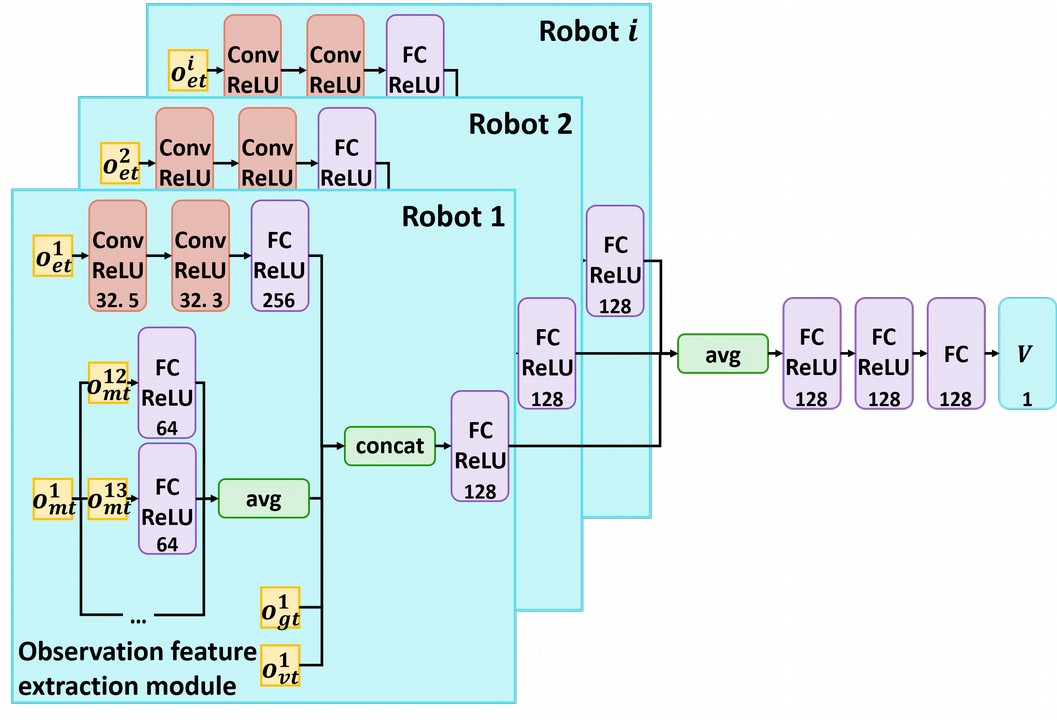}
  \vspace{-20pt}
  \caption{Value function network architecture.}
  \label{value function} 
  \vspace{-20pt}
\end{figure}
 
Though the value function $V_{\bm{\phi}}$ is centralized in the sense that it takes the global observation $\bm{o}_t$ of the robot team as input, it does not prohibit the decentralized execution since it is not used during execution. A neural network is also used as the non-linear function approximator for the value function $V_{\bm{\phi}}$. As shown in Fig. \ref{value function}, the architecture of value function network is similar to the policy network and consists of observation feature extraction modules $V_{{\bm{\phi}_1}}$ (wrapped by blue rectangles in Fig. \ref{value function}) and subsequent modules $V_{{\bm{\phi}_2}}$ (i.e., $\bm{\phi} = [\bm{\phi}_1, \bm{\phi}_2]$). Instead of sharing the whole subnetworks by all robots like the policy network, only the observation feature extraction module $V_{\bm{\phi}_1}$ is shared in the value function network. The insight of not sharing the subsequent modules $V_{\bm{\phi}_2}$ is that the expected accumulated reward depends on the global observations of the robot team. Since homogeneous robots are considered in this paper, the value function should be insensitive to the order of robots $(e.g., V_{\bm{\phi}}([\bm{o}^1_t,\bm{o}^2_t])=V_{\bm{\phi}}([\bm{o}^2_t,\bm{o}^1_t)])$. This property is guaranteed by sharing the observation feature extraction module among the robots as well as averaging extracted features before further processing.

\subsection{Optimization with proximal policy optimization}

\begin{figure}[!t]
	\removelatexerror
	\begin{algorithm}[H]
	\label{algo:ppo}
	\SetAlgoLined
	\BlankLine
	
	Initialize ${\bm{\theta}}_{old}$ and ${\bm{\phi}}_{old}$\;
	$\bm{\theta} \leftarrow \bm{\theta}_{old}, \bm{\phi} \leftarrow \bm{\phi}_{old}$ \;
	\For{iteration $= 1, 2, \dots, I$}{
		\tcp{Collect data in parallel}
		\For{worker $w = 1, 2, \dots, W$}{
			\For{time step $t = 1, 2, \dots, T$}{
				\For{robot $i = 1, 2, \dots, N$}{
					The $i$-th robot computes the action $\bm{a}^i_{wt}$ with the robot-level policy $\hat{\pi}_{{\bm{\theta}}_{old}}$ based on the local observation $\bm{o}^i_{wt}$\;
				}
				The robot team interacts with the environment with a team-level action $\bm{a}_{wt}$ by concatenating robot-level actions $\bm{a}^i_{wt}$ (i.e., $\bm{a}_{wt}=[\bm{a}^1_{wt}, \bm{a}^2_{wt}, \dots, \bm{a}^N_{wt}]$)\;
			}
			Collect trajectories\; 
		}
		\BlankLine
		
		\tcp{Update parameters}
		\For{epoch $e = 1, 2, \dots, E$}{
			\tcp{Update policy}
			Update $\bm{\theta}$ w.r.t the policy objective function $L_\pi(\bm{\theta})$ of PPO, which depends on the value function $V_{\bm{\phi}}$\;
			\tcp{Update value function}
			Update $\bm{\phi}$ w.r.t the value function objective function $L_V(\bm{\phi})$ of PPO\;
		}
		$\bm{\theta}_{old} \leftarrow \bm{\theta}$, $\bm{\phi}_{old} \leftarrow \bm{\phi}$ \;
		
	}
	\caption{Optimizing the decentralized policy with PPO}
	\end{algorithm}
	\vspace{-24pt}
\end{figure}
 
With the designed policy network and value function network, a decentralized policy $\hat{\pi}_{\bm{\theta}}$ can be derived in a centralized learning and decentralized executing paradigm once an actor-critic-based DRL algorithm is applied. In this paper, the recently proposed proximal policy optimization (PPO) \cite{schulman2017proximal}, which is a state-of-the-art DRL algorithm and guarantees stable performance improvement during training.

As summarized in Algorithm \ref{algo:ppo}, the optimization process alternates collecting data and updating parameters. In the data collection stage, multiple workers (a worker refers to a robot team in our case) interact with the environment in parallel and collect a batch of trajectories, which are used in the update step. In the parameter update procedure, the parameters $\bm{\theta}$ and $\bm{\phi}$ are optimized with respect to the objective functions of PPO $L_\pi(\bm{\theta})$ and $L_V(\bm{\phi})$ respectively.

 \section{SIMULATIONS AND EXPERIMENTS}
\label{sec:experiment}

In this section, simulations and real-world experiments on a team of three robots are carried out to illustrate the effectiveness of the proposed method.

\subsection{Training configurations}

The interactive simulation environment for training is implemented based on the OpenAI Gym \cite{1606.01540}. An open-source version of the PPO \cite{pytorchrl} based on Pytorch \cite{paszke2017automatic} is adapted to optimize the policies. All the programs run on a computer with Intel(R) Xeon(R) E5-2620 v4 CPU and Nvidia TITAN X (Pascal) GPU. The on-line running of the policy takes about $50$ ms on CPU and $5$ ms on GPU. The on-line computational complexity is low and efficient for real-time implementations. Detailed training settings are defined as follows: 

\begin{itemize}
\item Robot team settings: robot number $N=3$, robot radius $R=0.5 \text{m}$, and robot maximum velocity $v_{max} = 0.7 \text{m/s}$. Each robot is equipped with $30$ laser ejectors (i.e., a laser ejector every $12^\circ$) whose sensing range is $2$m. All robots are holonomic.
\item Environment settings: environment size is $10 \text{m} \times 10 \text{m}$. Obstacles of different sizes and shapes are generated randomly (look like scenarios in Fig. \ref{random}). Time interval $\Delta t=0.5\text{s}$. In each run, the positions of the robot team and the goal are randomly initialized.
\item Reward settings: efficiency term $r_e = -0.5$, weights of reward functions $w_g = 10, w_c = 1, w_f = 10, w_p = 5$, goal reward $r_{goal} = 10$, collision penalty $r_{collision} = -100$, radius of goal zone $\epsilon_g = 0.15\text{m}$, and communication range threshold $d=3.5\text{m}$.
\item PPO settings: number of iteration $I=24000$, epoch $E=4$, number of workers $W=16$, and maximum time step $T=128$. Adam optimizer \cite{kingma2014adam} is used to optimize parameters $\bm{\theta}$ and $\bm{\phi}$. The learning rates ($lr_\theta$ and $lr_\phi$) in the first half and the second half of the training procedure are set to $2.5 \times 10^{-4}$ and $5.0 \times 10^{-5}$ respectively.
\end{itemize}

\subsection{Simulation results on various scenarios}

We verify the effectiveness of our proposed method by applying the learned policy in different scenarios.

\begin{figure}
  \centering
  \subfloat[Block.]
  {  \setlength{\fboxsep}{0.8pt}\fbox{\includegraphics[width=0.3\linewidth]{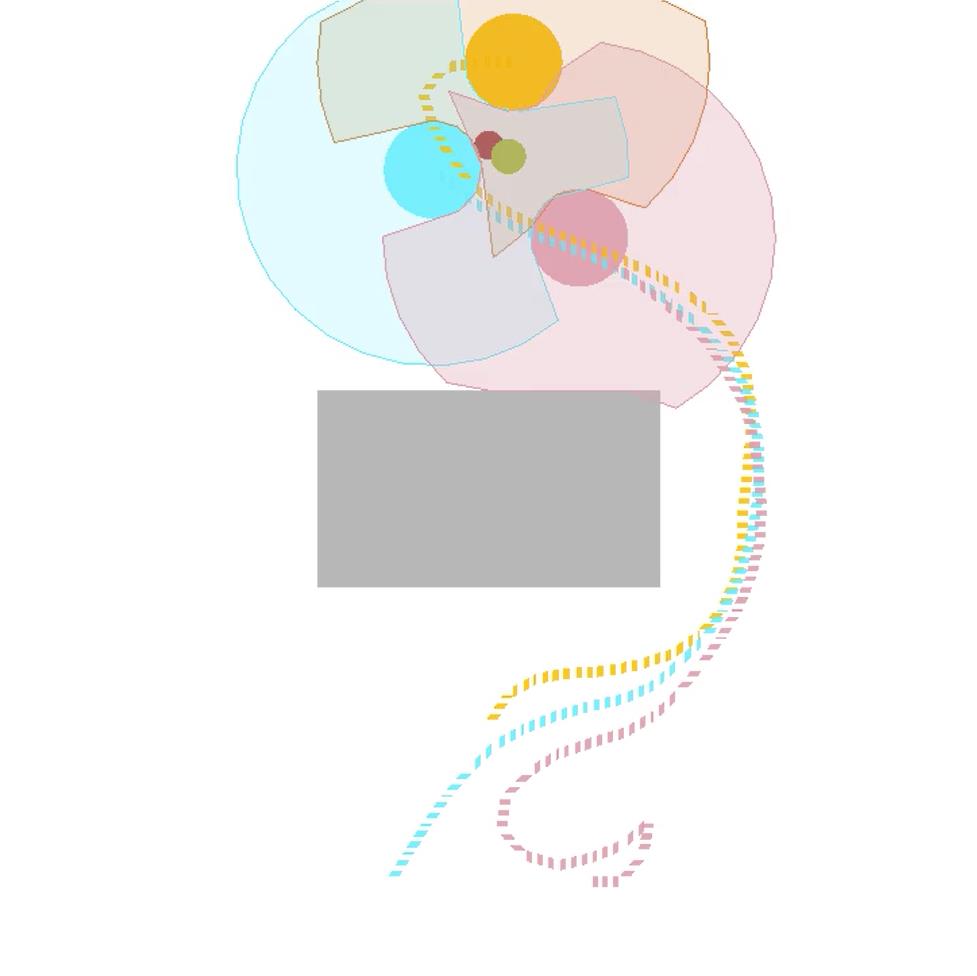}}
  \label{block}
  }
  \subfloat[Gear.]
  {  \setlength{\fboxsep}{0.8pt}\fbox{\includegraphics[width=0.3\linewidth]{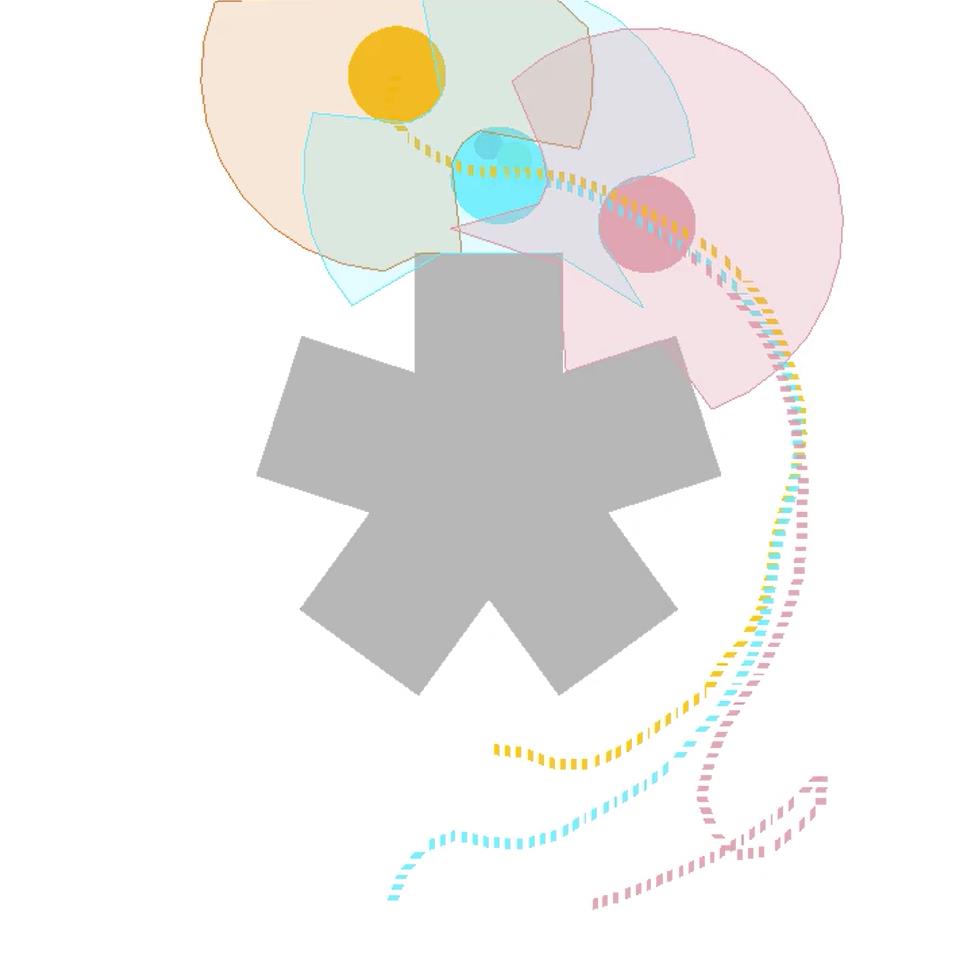}}
  \label{gear}
  }
  \subfloat[Groove.]
  {  \setlength{\fboxsep}{0.8pt}\fbox{\includegraphics[width=0.3\linewidth]{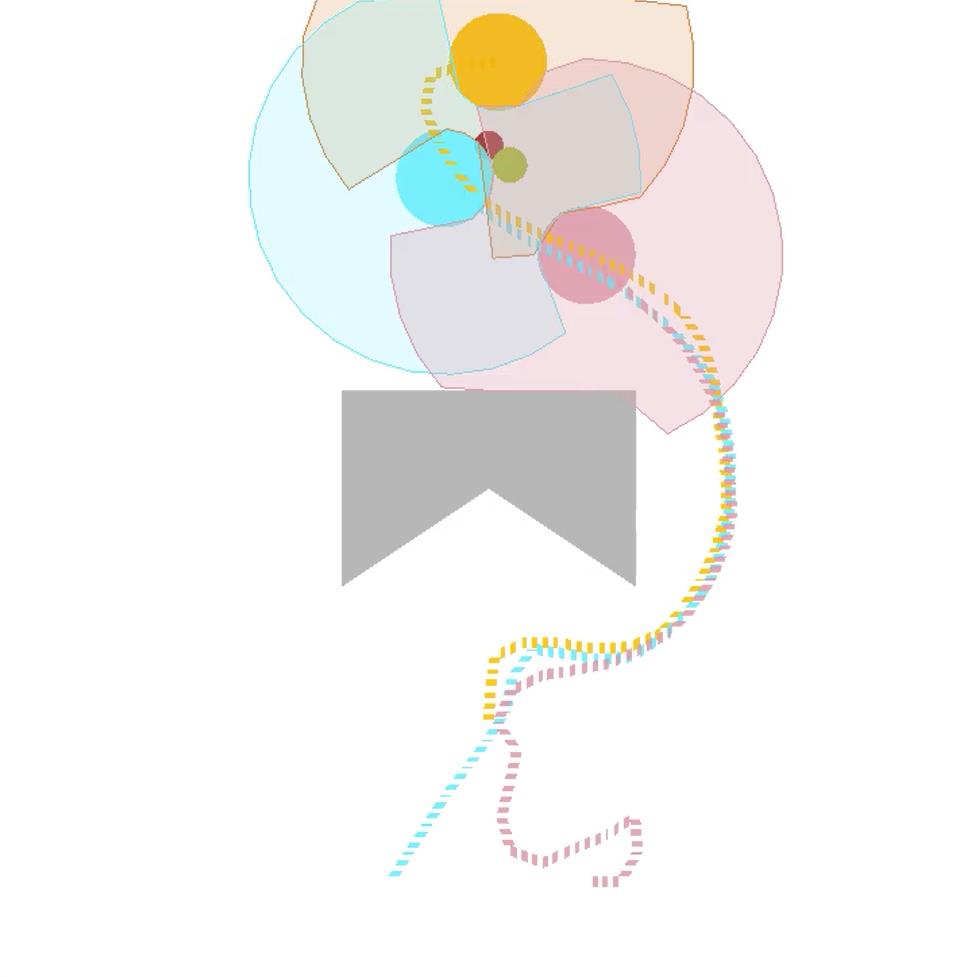}}
  \label{groove} 
  }
  \caption{Avoid obstacles of different shapes. Robots are pink, brown and blue circles. Obstacles are grey polygons. The goal is the dark red circle. Shaded discs are lidar measurements. Green circle is the centroid of the robot team and colored dashed lines are the trajectories of robots with corresponding colors. The color scheme applies across all simulation results.}
  \label{obs_shape}
  \vspace{-20pt}
\end{figure} 
\subsubsection{Obstacles of different shapes}

When operating in an unknown complex environment, a multi-robot team may encounter obstacles of different shapes. Therefore, the capacity of handling the diversity of obstacle shape is an important metric to evaluate the performance of a navigation policy. In order to assess the ability of avoiding obstacles of different shapes, we set up scenarios where obstacles of different shapes lie in between the robot team and the goal. In particular, test cases cover both convex and concave obstacles including block, gear and groove. As shown in Fig. \ref{obs_shape}, the policy succeeds in all test cases, which implies the policy's capacity of avoiding obstacles of different shapes.

\subsubsection{Narrow scenarios}

\begin{figure}
  \centering
  \subfloat[Narrow passageway.]
  {  \setlength{\fboxsep}{0.8pt}\fbox{\includegraphics[width=0.4\linewidth]{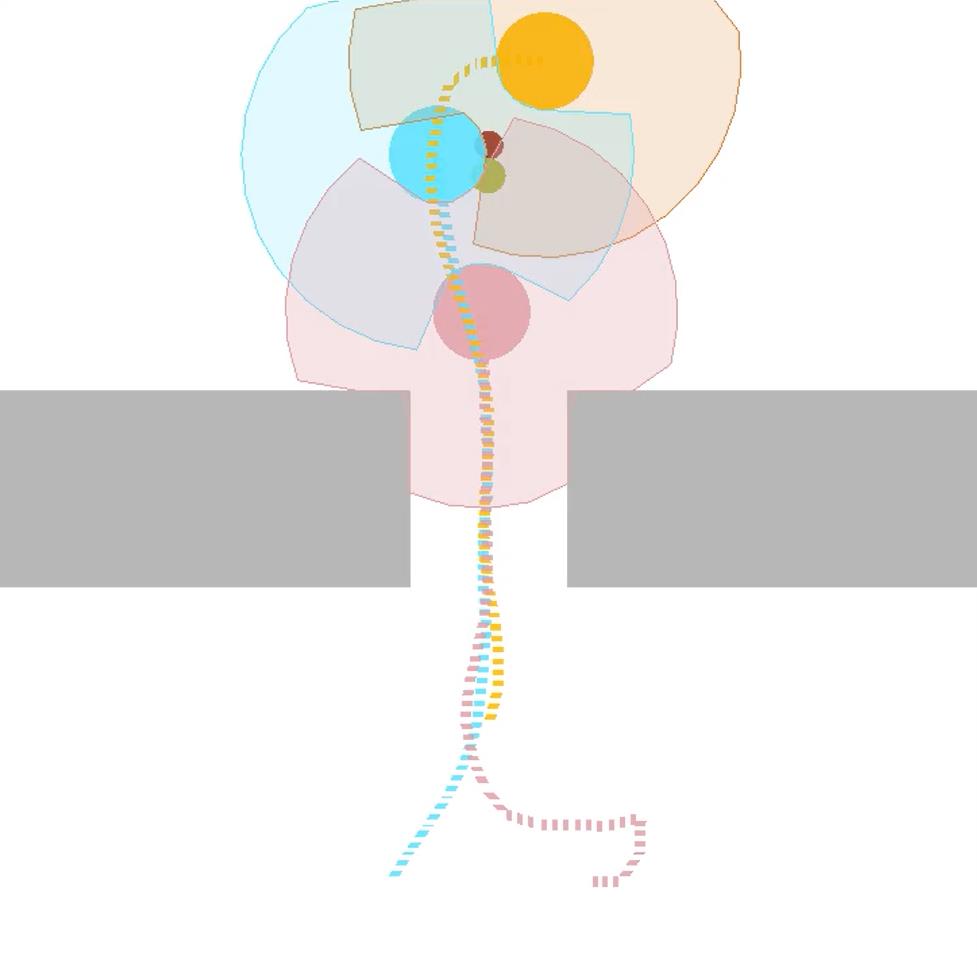}}
  \label{long_corridor}
  }
  \subfloat[Corridor with corners.]
  {  \setlength{\fboxsep}{0.8pt}\fbox{\includegraphics[width=0.4\linewidth]{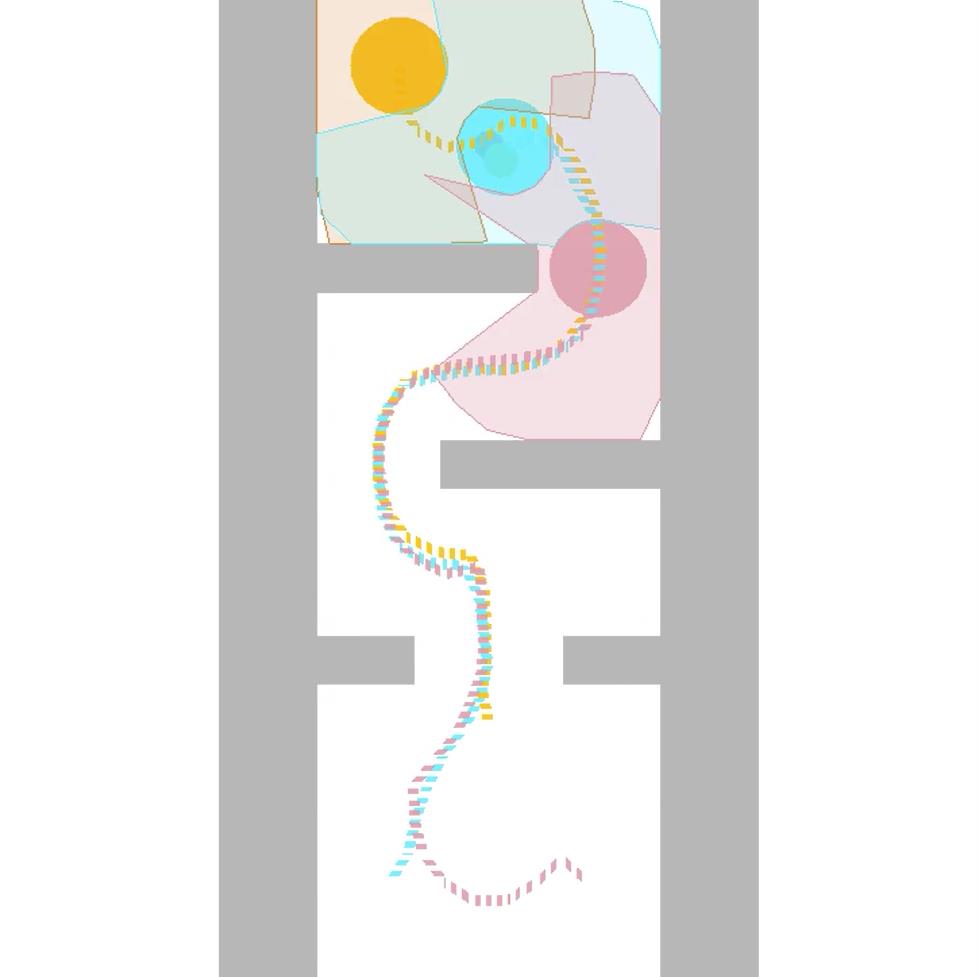}}
  \label{corridor_with_corners}
  }
  \caption{Navigation in narrow scenarios.}
  \label{corridor}
  \vspace{-20pt}
\end{figure} 
Narrow scenes are common in real world and is challenging in the sense that it imposes strict spacial constraints to the robot team. To verify the effectiveness of the derived policy in narrow scenes, we set up scenarios including narrow passageway and corridor with corners. It can be seen in the Fig. \ref{corridor} that the policy successfully navigates the centroid of the robot team to the goal in different narrow scenarios.

\subsubsection{Random scenarios}

\begin{figure}
  \centering
  \subfloat[Random scenario 1.]
  {  \setlength{\fboxsep}{0.8pt}\fbox{\includegraphics[width=0.4\linewidth]{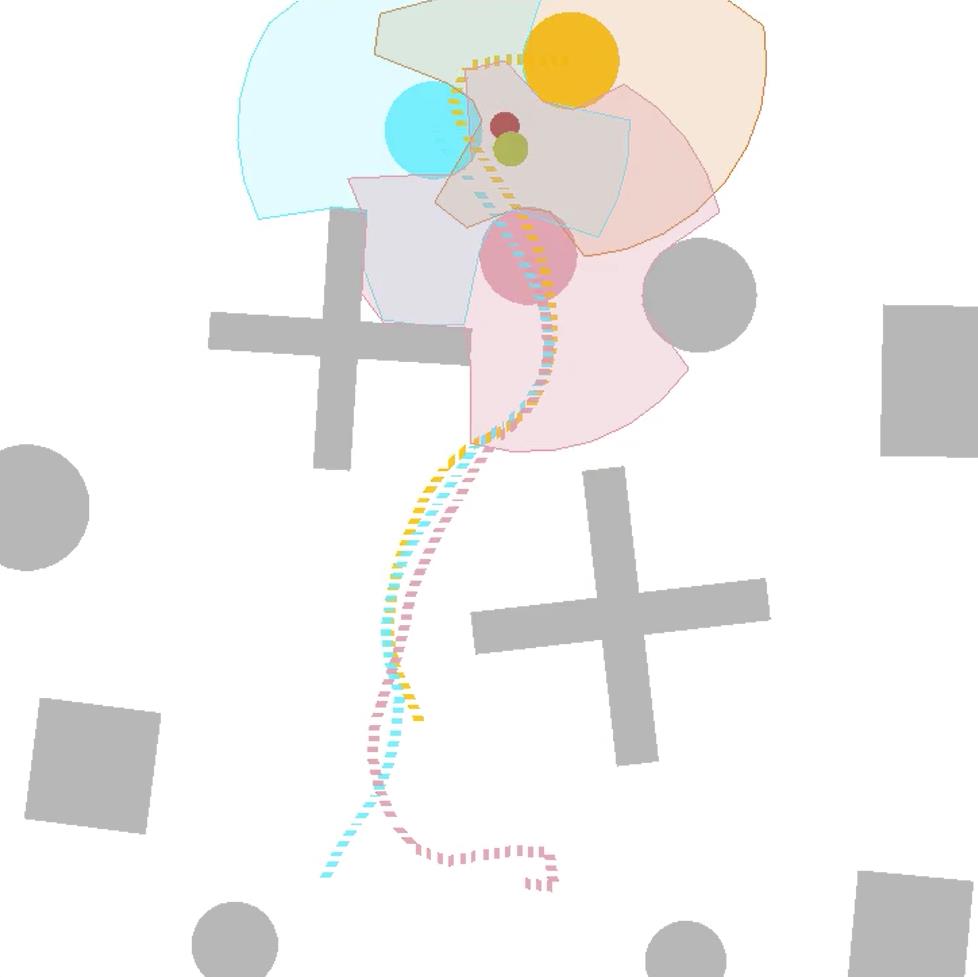}}
  \label{random1}
  }
  \subfloat[Random scenario 2.]
  {  \setlength{\fboxsep}{0.8pt}\fbox{\includegraphics[width=0.4\linewidth]{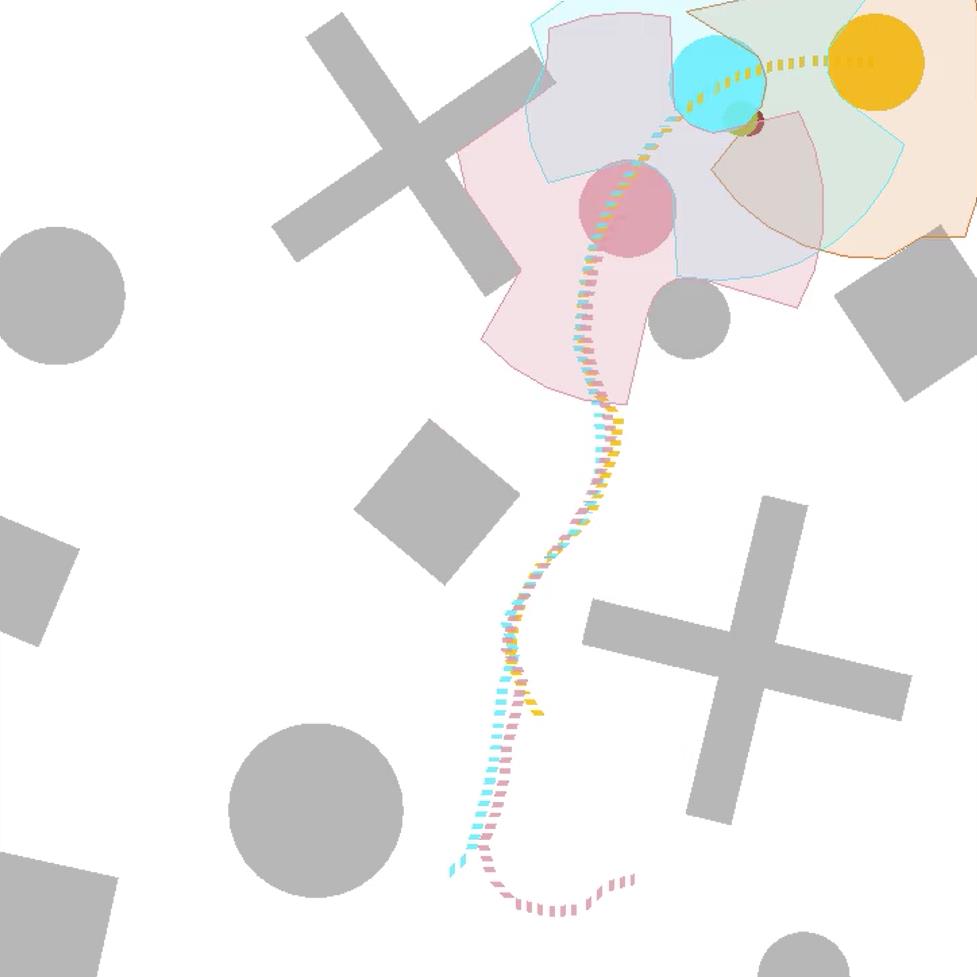}}
  \label{random2}
  }

  \caption{Navigation in random scenarios.}
  \label{random}
  \vspace{-20pt}
\end{figure}

Random scenarios are frequently used to evaluate the performance of a navigation policy. We randomly generate scenarios with obstacles of different shapes and sizes to test the derived policy. As shown in Fig. \ref{random}, the policy successfully navigates the centroid of the robot team to the goal in random scenarios.

Simulation results show that our method can navigate the robot team through various complex scenarios. Note that our policy is run in a decentralized manner and computes continuous velocity control commands with raw laser measurements, which implies that our policy can achieve end-to-end decentralized cooperation in difficult environments.

\subsection{A comparison with a rule-based method}

\begin{figure}
  \centering
  \subfloat[Our method.]
  {  \setlength{\fboxsep}{0.8pt}\fbox{\includegraphics[width=0.4\linewidth]{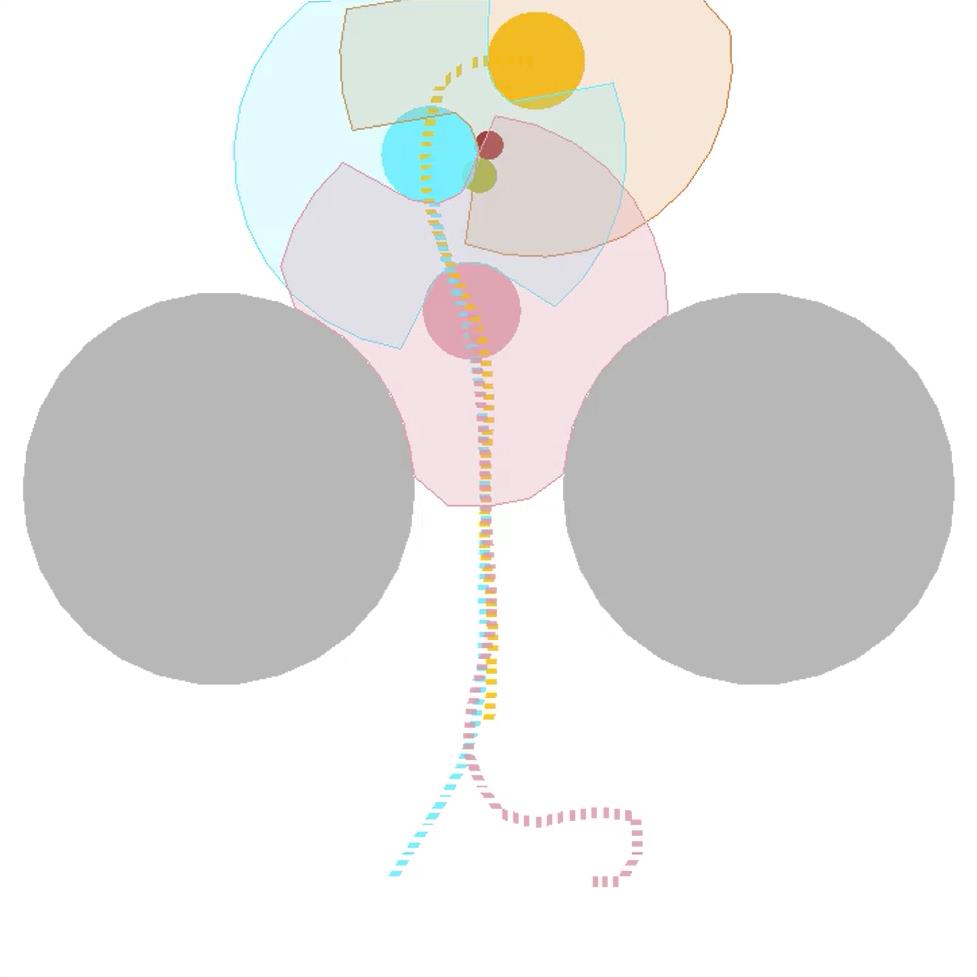}}
  \label{apf_rl}
  }
  \subfloat[APF-based method.]
  {  \setlength{\fboxsep}{0.8pt}\fbox{\includegraphics[width=0.4\linewidth]{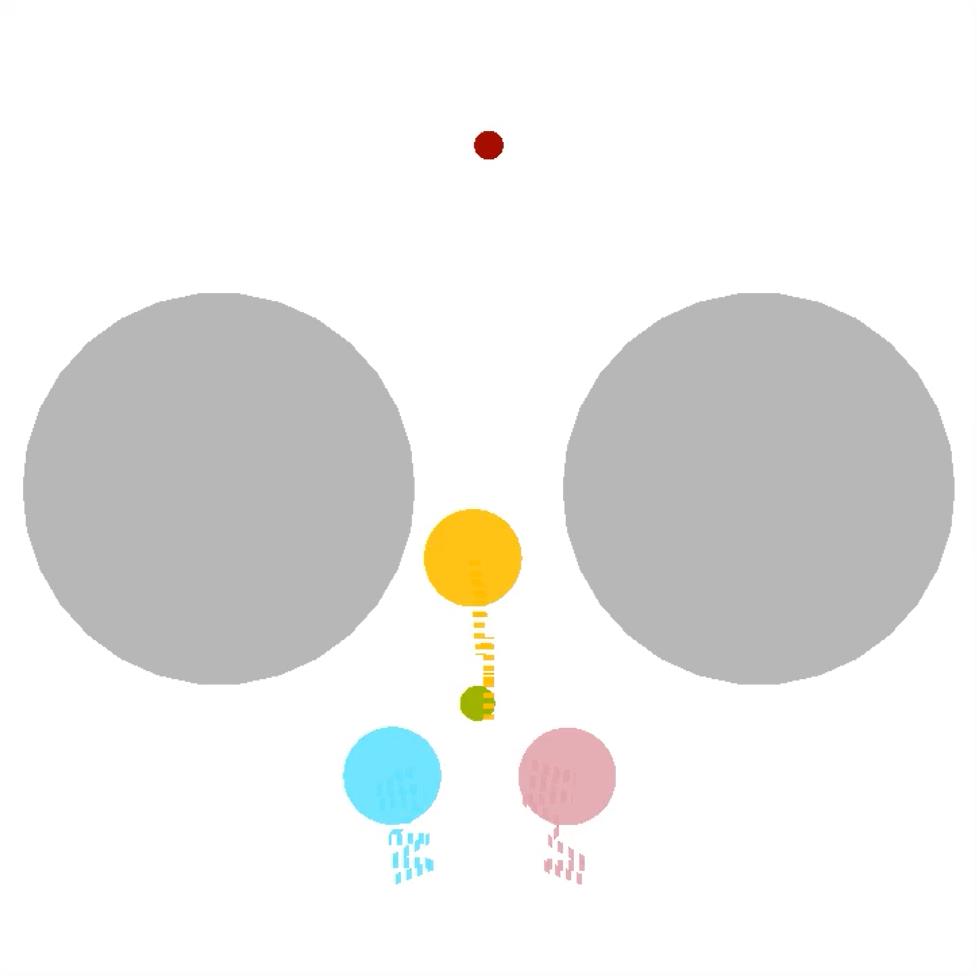}}
  \label{apf_apf} 
  }
  \caption{A comparison with an APF-based method.}
  \label{apf}
  \vspace{-20pt}
\end{figure}

A comparison with a rule-based method (in particular, an artificial potential field (APF)-based method \cite{balch2000social}) is conducted to investigate the performance of the proposed method. As shown in Fig. \ref{apf}, our policy navigates the robot team through the narrow passageway, while the APF-based method gets stuck at the entrance since the repulsive forces from obstacles neutralize the attractive force from the goal. Note that in this experiment, our policy computes velocity control commands merely based on local observations while the APF-based method can obtain global environment information.

\subsection{A comparison with other decentralization mechanism}

\begin{figure} 
  \centering
  \subfloat[Convergence.]
  {  \includegraphics[width=0.47\linewidth]{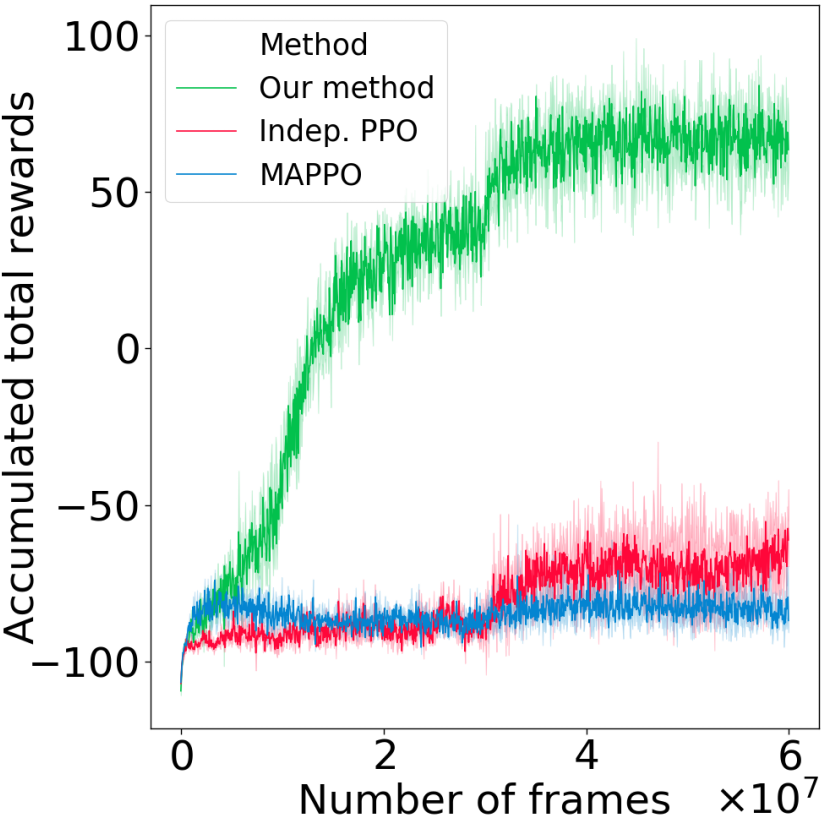}
  \label{metric:convergence} 
  }
  \subfloat[Success rate.]
  {  \includegraphics[width=0.47\linewidth]{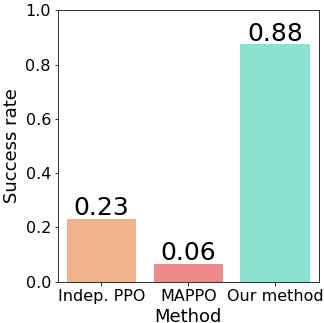}
  \label{metric:success_rate}
  }
  \vspace{-12pt}
  \\
  \subfloat[Connectivity.]
  {  \includegraphics[width=0.47\linewidth]{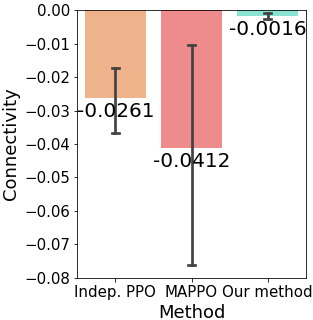}
  \label{metric:connectivity}
  }
  \subfloat[Efficiency.]
  {  \includegraphics[width=0.47\linewidth]{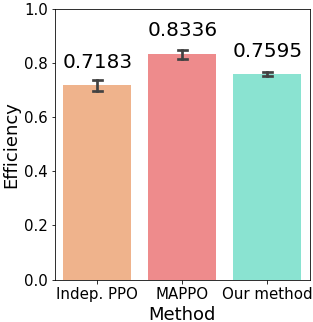}
  \label{metric:efficiency}
  }
  \hfill
  \vspace{-5pt}
  \caption{A quantitative comparison among independent PPO, MAPPO and our method.}
  \label{metric}
  \vspace{-20pt}
\end{figure}
 
To verify the effectiveness of the adopted mechanism for centralized learning and decentralized execution, we conduct quantitative comparisons with ablated variants of our method called independent PPO (which ablates the centralized learning procedure) and MAPPO (which ablates the joint policy update). In particular, the independent PPO follows the paradigm of independent Q-learning \cite{tan1993multi}, where each robot independently updates the policy and treats other robots as a part of the environment. It implies that the value function takes the local observation $\bm{o}^i_t$ as input. The MAPPO applies the mechanism proposed in MADDPG \cite{NIPS2017_7217} where value function takes the global observation $\bm{o}_t$ as input while robot-level policy $\hat{\pi}_{\bm{\theta}}$ for each agent is updated separately. In comparison, our method exploits the centralized learning and updates the team-level joint policy $\pi_{\bm{\theta}}$ for the robot team.

The following quantitative metrics are adopted to evaluate the performance:

\begin{itemize}
\item Success rate: The ratio of the success times $n_{success}$ over the number of the total test cases $n_{total}$, i.e., $n_{success}/n_{total}$.
\item Connectivity: The negative average violation of the distance constraint over a trajectory, i.e., $-(1 / T^\prime) \sum_{t=1}^{T^\prime} c_{t}$, where $c_{t} = \avg(\{ \max(0, \| \bm{p}_{t}^i - \bm{p}_{t}^j  \|_2 - d) \ \vert \ \forall i \neq j \})$ is the average violation of distance constrain at time step $t$ and $T^\prime$ is the trajectory length.
\item Efficiency: The ratio of the lower bound of travel time $t_{lb}$ (dividing Euclidean distance from the initial position to the goal by the maximum velocity) over the actual travel time $t_{travel}$, i.e., $t_{lb} / t_{travel} $.
\end{itemize}

For all quantitative metrics, higher values indicate better performance. For efficiency and connectivity, we only consider successful navigations and neglect failure cases. To alleviate random errors, 3 policies are trained with different random seeds for each method. The final quantitative metrics of a method is the average performance of 3 derived policies.

100 random scenarios (look like scenarios in Fig. \ref{random}) are generated to evaluate policies derived by independent PPO, MAPPO and our method. The curves of accumulated total rewards over training process are shown in Fig. \ref{metric:convergence}. It can be seen that our method converges to much higher accumulated total rewards in comparison with independent PPO and MAPPO. As for quantitative metrics, our method significantly outperforms independent PPO and MAPPO in success rate (Fig. \ref{metric:success_rate}) and connectivity (Fig. \ref{metric:connectivity}). It can be seen from Fig. \ref{metric:efficiency} that the MAPPO slightly surpass our method in efficiency, while the performance of the former severely degenerates with respect to the success rate and connectivity. From the quantitative results, we can conclude that both centralized learning and joint policy update are essential for the navigation task, which highly demands group behaviors.

\begin{figure} 
  \centering
  {
  \setlength{\fboxsep}{0.0pt}\fbox{\includegraphics[width=0.5\linewidth]{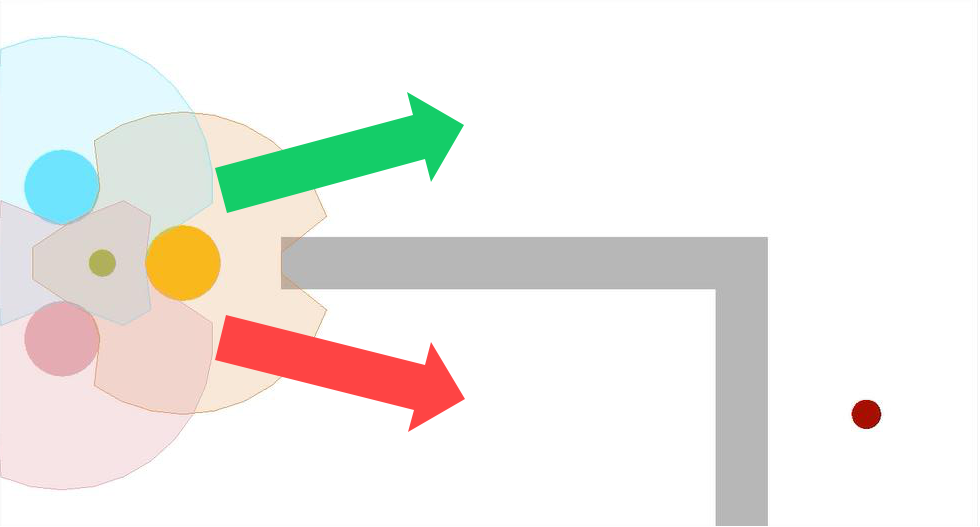}}
  }
  \vspace{-5pt}
  \caption{A demonstration of typical failure cases. The robot team cannot predict the dead end merely based on local observations. Therefore, the robot team may choose to follow the red arrow so as to reach the goal as fast as possible and runs into the dead end.}
  \label{failure} 
  \vspace{-21.64pt}
\end{figure}

It's noticed that the policy derived by our method may fail in some of the navigation tasks. A typical reason for the failures can be demonstrated by Fig. \ref{failure}. When the robot team encounters a crossway and has to decide which way to go (following the green arrow or the red arrow), it's difficult for the policy to predict the dead end beforehand merely based on local observations. Therefore, trying to approach the goal directly (following the red arrow) is a reasonable choice, which is also a behavior (reaching the goal as fast as possible) encouraged by the reward function (\ref{reward_function}). Consequently, the robot team runs into the dead end and fails the navigation. Our method currently relies on guide points from higher level path planning algorithm to avoid such failures.

\begin{figure}
  \centering
  \subfloat[The chasis of an UGV with an onboard computer (Nvidia Jetson TX2).]
  {  \includegraphics[width=0.6\linewidth]{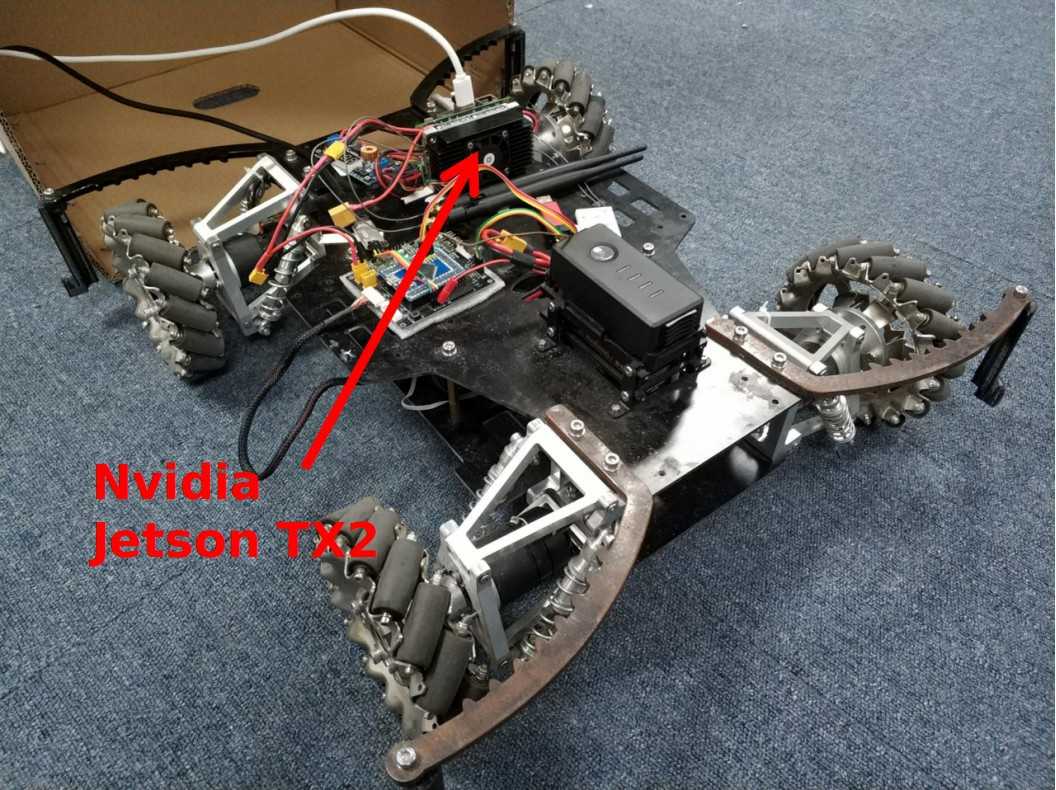}
  \label{chasis}
  }
  \vspace{-10pt}
  \\
  \subfloat[A holonomic UGV equipped with a 2D laser scanner (RPLIDAR-A2).]
  {  \includegraphics[width=0.48\linewidth]{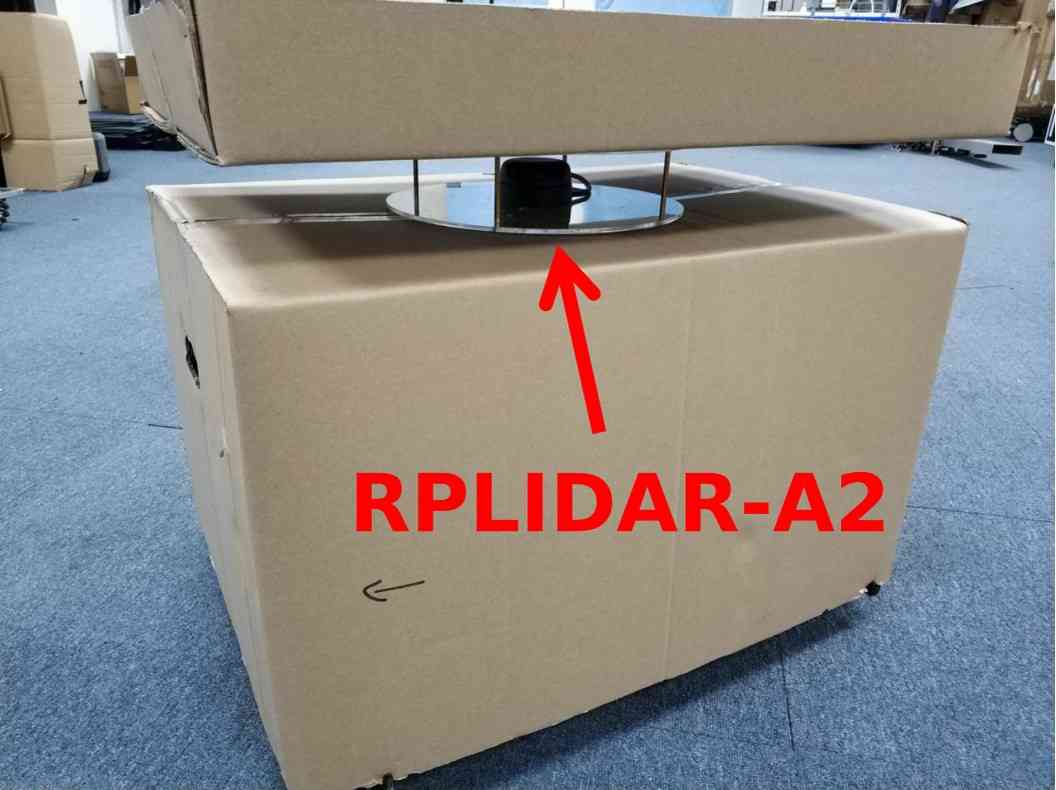}
  \label{ugv}
  }
%  \hfill
  \subfloat[A robot team of three holonomic UGVs.]
  {  \includegraphics[width=0.48\linewidth]{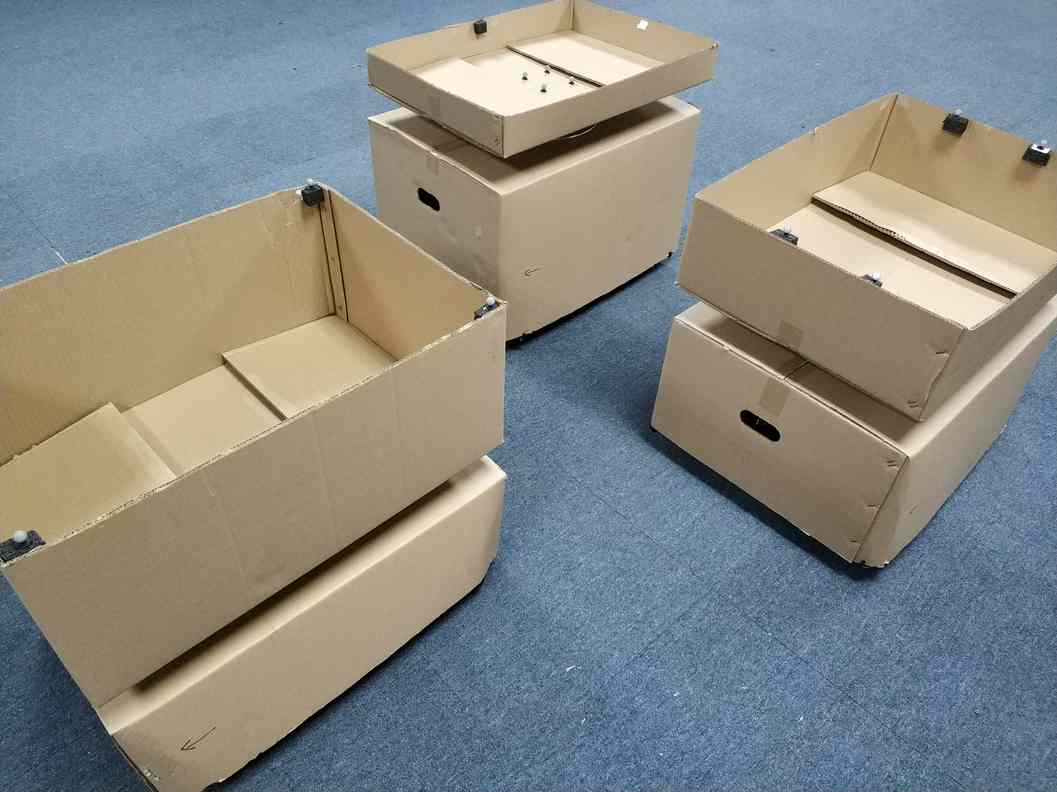}
  \label{team}
  }
  \caption{Real-world experimental platform.}
  \label{platform} 
  \vspace{-22pt}
\end{figure}
 
\begin{figure*}
  \centering
  \subfloat[Initial position of the robot team and obstacles.]
  {  \hfill
  \includegraphics[width=0.33\linewidth]{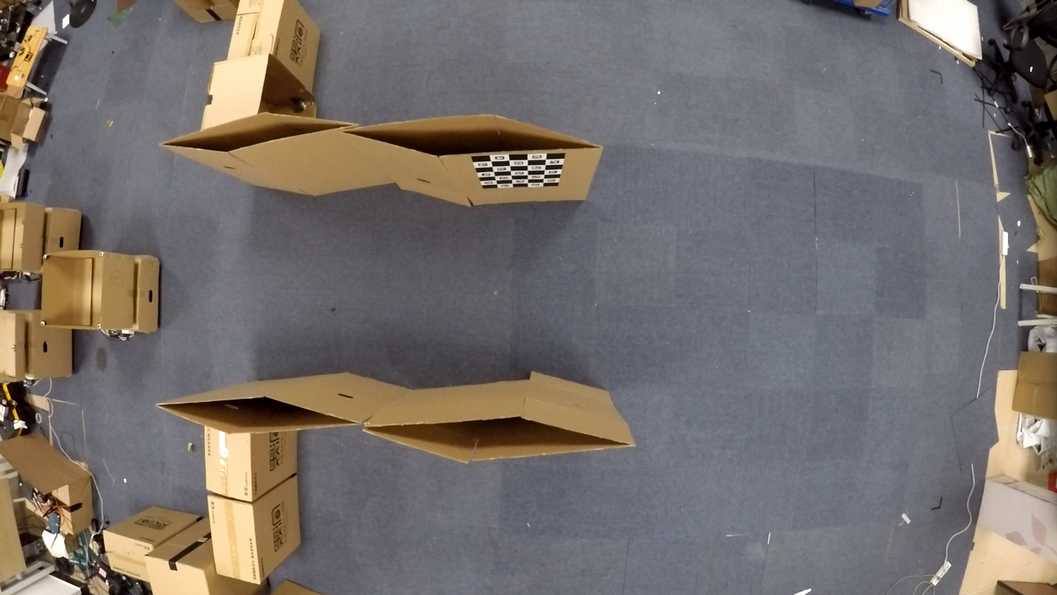}
  \label{interference:success rate}
  }
  \subfloat[The robot team switches to line formation.]
  {  \hfill
  \includegraphics[width=0.33\linewidth]{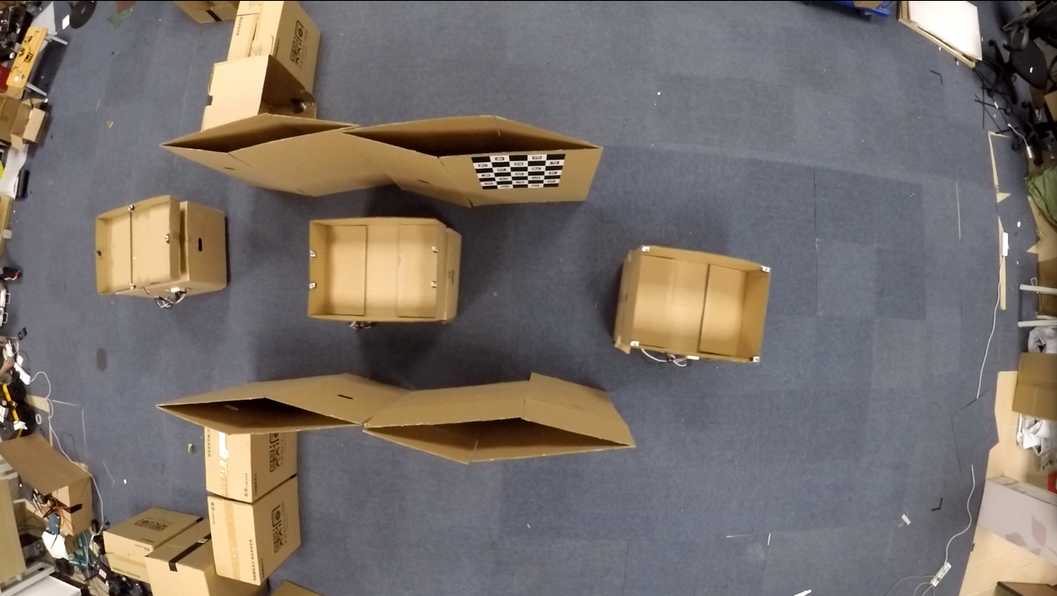}
  \label{interference:connectivity}
  }
  \subfloat[The robot team reaches the goal.]
  {  \includegraphics[width=0.33\linewidth]{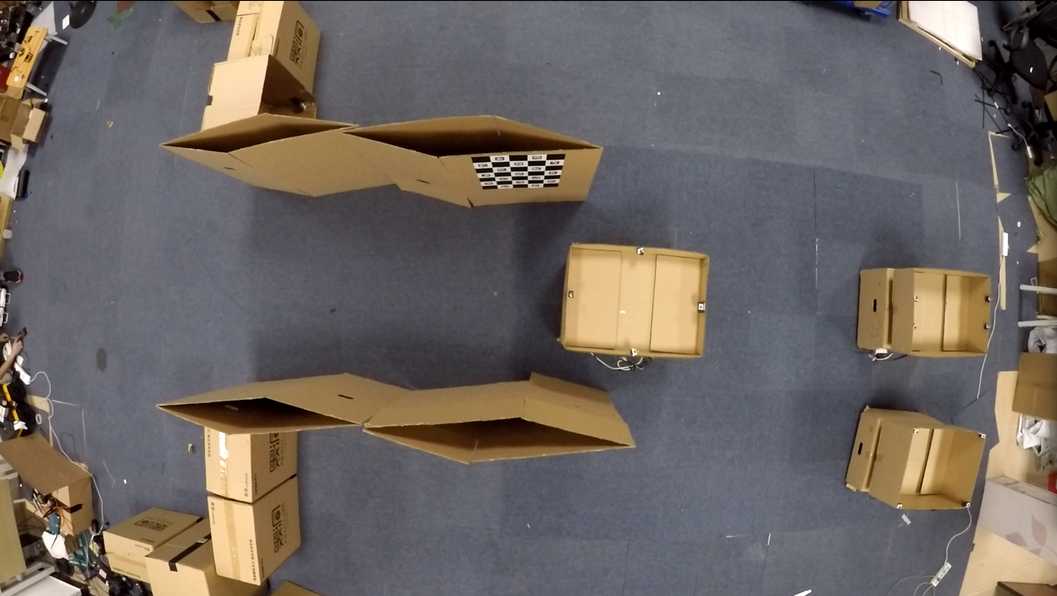}
  \label{interference:smoothness} 
  }
  \hfill
  \caption{Snapshots of a team of 3 holonomic UGVs navigating through obstacles.}
  \label{interference} 
  \vspace{-20pt}
\end{figure*}
 
\subsection{Real-world experimental results}

Real-world experiments are performed to verify the practicability of the proposed method. Three holonomic UGVs (as shown in Fig. \ref{platform}) are built to form a robot team, which serves as a platform to test derived policies. In order to enable real-time end-to-end decentralized execution, each UGV is equipped with a 2D-laser scanner (RPLIDAR-A2) and an onboard computer (Nvidia Jetson TX2). We wrap the UGVs with boxes and set laser scanners at different heights to make laser scanners detect teammate UGVs. The OptiTrack motion capture system is used to provide the positions of the robots.
 
The snapshots of an experiment are shown in Fig. \ref{interference}. In this experiment, a narrow corridor lies between the robot team (which is initialized on the left) and the goal (which is set on the right). It can be seen that the derived policy switches the robot team into line formation to go through the narrow corridor and successfully navigates the centroid of the robot team to the goal. Note that the policy is run in a decentralized manner and directly maps raw laser measurements into continuous velocities without constructing obstacle maps.

We think the success of sim-to-real deployment partially credits to the input consistency. In particular, the inputs (e.g., laser measurements) of the policy are relatively consistent in simulations and real-world environments (compared with RGB inputs which are heavily affected by light). Moreover, scene pattern similarity between simulations and real-world experiments is also an important factor for the success. In spite of the consistency and similarity, it should be pointed out that the sim-to-real deployment is challenging because both the observations and actions in real world are fairly noisy and imperfect in comparison with simulations. The success of real-world experiments shows the robustness and practicability of the proposed method.

A demo video of the multi-robot navigation can be seen in https://youtu.be/G8UZU1hk9K4.
 \section{CONCLUSIONS}
\label{sec:conclusion}

This paper proposes a DRL-based method to derive decentralized policies for a robot team to navigate through unknown complex environments safely while maintaining connectivity. The derived policy directly maps raw laser measurements to continuous velocity control commands without constructing the obstacle maps. By means of the learned policies, the robots cooperatively plan the motions to accomplish the navigation task of the robot team using each robot's local observations. Simulation results in various scenarios verify the effectiveness of the proposed method. Furthermore, indoor real-world experimental results demonstrate that the three-robot team can navigate through the dense obstacles. In the future work, the learning-based navigation approach will be further studied in robot teams of larger scale.

 \section*{ACKNOWLEDGMENT}

This work is supported by Major Program of Science and Technology Planning Project of Guangdong Province (2017B010116003), NSFC-Shenzhen Robotics Projects (U1613211), and Guangdong Natural Science Foundation (1614050001452, 2017A030310050). 
                                                                                                                                                                          
\bibliographystyle{IEEEtran}
\bibliography{IEEEabrv,root}

\end{document}